\documentclass[11pt]{article}

\usepackage{fullpage}
\usepackage{mathptmx}
\usepackage{times}
\usepackage{microtype}
\usepackage{graphicx}
\usepackage{subfigure}
\usepackage{hyperref}
\usepackage{multicol}

\usepackage{amsmath,amsbsy,amsfonts,amssymb,amsthm}
\usepackage{algorithmic,mathtools}
\usepackage{mathtools}
\usepackage{color}
\usepackage{xspace}
\usepackage{wrapfig}

\DeclareMathOperator*{\argmin}{arg\,min}

\newcommand{\norm}[1]{\left\lVert#1\right\rVert}

\newcommand{\eps}{\varepsilon}
\newcommand{\comp}{\textsc{cr}\xspace}
\newcommand{\emd}{\textsc{emd}\xspace}

\newcommand{\ex}{\mathbb{E}}
\newcommand{\pr}{\mathbb{P}}

\newcommand{\eat}[1]{}

\usepackage{xcolor}

\newtheorem{theorem}{Theorem}
\newtheorem*{theorem*}{Theorem}
\newtheorem{lemma}[theorem]{Lemma}
\newtheorem*{claim*}{Claim}
\newtheorem{corollary}[theorem]{Corollary}
\newtheorem*{corollary*}{Corollary}
\newtheorem{assumption}{Assumption}
\newtheorem{definition}{Definition}
\newcommand{\ml}{{\sc ml}\xspace}
\usepackage{algorithm}

\title{Customizing ML Predictions for Online Algorithms}
\author{Keerti Anand\thanks{Department of Computer Science, Duke University, Durham, NC, USA. Emails: {\tt \{kanand, rongge, debmalya\}@cs.duke.edu}.}
\and Rong Ge\footnotemark[1]
\and Debmalya Panigrahi\footnotemark[1]
}
\date{}

\begin{document}
\maketitle
\begin{abstract}
%    Traditionally, online algorithms optimize the worst-case competitive ratio between the algorithm and the optimal solution. To overcome the inherent pessimism of worst-case analysis, 
    A popular line of recent research incorporates \ml advice in the design of online algorithms to improve their performance in typical instances. These papers treat the \ml algorithm as a black-box, and redesign online algorithms to take advantage of \ml predictions. In this paper, we ask the complementary question: can we redesign \ml algorithms to provide better predictions for online algorithms? We explore this question in the context of the classic rent-or-buy problem, and show that incorporating optimization benchmarks in \ml loss functions leads to significantly better performance, while maintaining a worst-case adversarial result when the advice is completely wrong. We support this finding both through theoretical bounds and numerical simulations.
    %, and posit that ``learning for optimization'' is a fertile area for future research.
\end{abstract}

\clearpage

\section{Introduction}
\label{sec:introduction}
Optimization under uncertainty is a classic theme in the fields of algorithm design and machine learning. In the former, the framework of online algorithms adopts a conservative approach and optimizes for the worst case (or adversarial) future. While this ensures robustness, the inherent pessimism of the adversarial approach  often results in weak guarantees. Machine learning (\ml), on the other hand, takes a more optimistic approach of trying to predict the future by fitting an appropriate model to past data. Indeed, a popular line of recent research is to incorporate \ml predictions in the design of online algorithms to improve their performance while preserving the inherent robustness of the framework (see related work for references). In this line of research, \ml is used as a {\em black box}, and the focus is on re-designing online algorithms to use predictions generated by any \ml technique. In this paper, we ask the complementary question: {\em can we re-design learning algorithms to better serve optimization objectives?}

The key to this question is the observation that unlike in a generic learning setting, we are not interested in traditional loss functions such as classification error or mean-squared loss, but only in the eventual performance of the online algorithm.
The performance of the online algorithm is measured by its {\em competitive ratio} \--- the worst-case ratio between the cost of the online algorithm's solution and that of the (offline) optimum. By leveraging \ml predictions, one can hope to achieve a better competitive ratio in the typical case. Even if the \ml algorithm does not make accurate predictions, it suffices if the learning errors do not adversely affect the decisions taken by the online algorithm. Instead of treating the learning algorithm and the subsequent optimization as independent modules as in the previous line of work, we ask if we can improve the overall online algorithm by designing them in conjunction. That is, we seek to design a learning algorithm specific to the optimization task at hand, and an optimization algorithm that is aware of the learning algorithm that generated the predictions.

We investigate this question in the context of the classic {\em rent-or-buy} (a.k.a. {\em ski rental}) problem. In this problem, the algorithm is faced with one of two choices: a small recurring (rental) cost, or a large (buying) cost that has to be paid once but no cost thereafter. This choice routinely arises in our daily lives such as in the decision to rent or buy a house, corporate decisions to rent or buy data centers, expensive equipment, and so on. Naturally, the optimal choice depends on the duration of use, a longer duration justifying the decision to buy instead of renting. But, this is where the uncertainty lies: the length of use is often not known in advance. The ski rental problem is perhaps the most fundamental, and structurally simplest, of all problems in online algorithms, and has been widely studied in many contexts (see, e.g., \cite{KMMO94,KKR03,LPR08, KKP13,K14}), including that of online algorithms with \ml predictions~\cite{purohit2018improving,GollapudiP19}. We formally define this problem next.

\smallskip\noindent{\bf The ski rental problem.}
In the ski rental problem, a skier has two options: to buy skis at a one time cost of \$$B$ or to rent them at a cost of \$$1$ per day. The skier does not know the length of the ski season in advance, and only learns it once the season ends. Note that if the length of the season were known, then the optimal policy is to buy at the beginning of the season if it lasts longer than $B$ days, and rent every day if it is shorter. But, in the absence of this information, an algorithm has to decide the duration of renting skis before buying them. It is well-known that the best competitive ratio achievable by a deterministic algorithm for this problem is $2$ (e.g., \cite{KMRS88}), and that by a randomized algorithm is $\frac{e}{e-1}$ (e.g., \cite{KMMO94}). The ski-rental problem~\cite{KMMO94, LPR08, KKP13,K14}, and variants such as TCP acknowledgment~\cite{KKR03}, the parking permit problem~\cite{M05}, snoopy caching~\cite{KMRS88}, etc. model the fundamental difficulty in decision making under uncertainty in many situations.

%We investigate the problem of Ski Rental (which is a general name to a class of online problems where we face the choice between continuing to pay a recurring cost for a certain resource (rent) or pay a fixed one time cost (buy) for the resource at some point which eliminates the future recurring costs.

%In the Ski Rental Problem, the player is going skiing : but the total number of days she is going to ski is not known in advance. She faces two choices: she can continue to rent the ski for the entire ski season (say at $ \$1/day$) or at some point decide to buy it (at say cost $ \$ B $). If the number of days were known in advance, then the solution is straightforward: Buy at the beginning if the ski season lasts greater than or equal to $B$ days, otherwise rent.

%The Ski Rental Problem admits a 2-competitive deterministic algorithm \cite{karlin1988competitive} and a $\frac{e}{e-1}$ randomized algorithm \cite{karlin1994competitive}. Both of these results are known to be tight.

\smallskip\noindent{\bf The learning framework.}
We use a classic PAC learning framework. Namely, the learning algorithm observes feature vectors $x\in \mathbb{R}^d$ comprising, e.g., weather predictions, skier history, etc. and aims to predict scalars $y\in \mathbb{R}^+$ denoting the length of the ski season.
%The variable $y$ depends on a parameter $x \in \mathbb{R}^d$. (this parameter $x$ is known as a feature and can denote the past values of $y$, the weather/ temperature etc). In this section, we are going to assume that 
We assume that $(x,y)$ belongs to an unknown joint distribution $\mathbb{K}$. 
The learning algorithm observes $n$ samples (the ``training set'') from $\mathbb{K}$. Typically, these samples would be used to train a model that maps feature vectors $x$ to predictions $\tilde{y} = f(x)$ that minimizes some loss function (e.g., mean squared error, hinge loss, etc.) defined on $\mathbb{K}$. In our problem, however, the goal is not to predict the unknown $y$, but rather to optimize the solution to the ski rental instance defined by $y$. Consequently, the learning algorithm skips $y$ altogether and outputs a solution to the optimization problem directly. For the ski rental problem, this amounts to defining a function $\theta(x)$ that maps the feature vector $x$ to the duration of renting skis. The expected competitive ratio is then given by the competitive ratio of this policy $\theta(x)$ defined on distribution $\mathbb{K}$. We call this a ``learning-to-rent'' algorithm.

\smallskip\noindent{\bf Our Contributions.}
Our goal is to design a learning-to-rent algorithm with an expected competitive ratio of $(1+\eps)$, and analyze the dependence of the number of samples $n$ on the value of $\eps$. Contrast this with online algorithms for this problem that can at best achieve a competitive ratio of $\frac{e}{e-1}$ (e.g., \cite{KMMO94}).
If the joint distribution $(x,y)$ is arbitrary, then one cannot hope to achieve a competitive ratio of $(1+\eps)$ since every sample may have a different $x$ and the conditional distributions $y|x$ may be unrelated for different values of $x$. However, it is natural to assume that the joint distribution on $(x,y)$ is {\bf Lipschitz} in the sense that nearby values of $x$ imply similar conditional distributions $y|x$. Our first contribution (Theorem~\ref{thm:lipschitz}) is to design a learning-to-rent algorithm whose competitive ratio is within a factor of $(1+\eps)$ of the best competitive ratio achievable for that distribution, under only the Lipschitz assumption. First, we discretize the domain of $x$ using an $\epsilon$-net. Then, for each cell in the $\epsilon$-net, we have one of two cases. Either, there are sufficiently many samples to estimate the conditional distribution $y|x$. Or, a baseline online algorithm can be used for the cell if it has very few samples. The dependence of the number of samples $n$ on the number of feature dimensions $d$ is exponential, which we show is indeed necessary (Theorem~\ref{thm:lipschitz_lowerbound}). 

Our next goal is to improve the dependence on $d$ since the number of features in a typical setting can be rather large, which would make the previous algorithm prohibitively expensive. To this end, we use a PAC learning approach to address the problem. Since the optimal ski rental policy exhibits threshold behavior (rent throughout if $y < B$ and buy at the outset if $y \geq B$), we treat the underlying learning problem as a classification task. In particular, we introduce an auxiliary binary variable $z$ that captures the two regimes for the optimal ski rental policy:
\begin{equation*}
    z = \left\{\begin{array}{ll} 1 & \text{if } y\geq B\\0 & \text{if } y < B\end{array}\right.
\end{equation*}
Our first result is that if $z$ belongs to a concept class that is $(\eps, \delta)$ PAC-learnable from $x$, then we can obtain a learning-to-rent algorithm that achieves a competitive ratio of $(1 + 2\sqrt{\epsilon})$ with probability $1-\delta$. This implies, for instance, that if there were a linear classifier for $z$, then the number of required training samples $n$ to obtain a $(1+\eps)$ competitive algorithm can be decreased from exponential to linear in $d$, specifically $O(d/\eps^2)$. 
% \alert{This is a little confusing, since this bound is for a competitive ratio of $1+2\sqrt{\epsilon}$ whereas we were previously talking about a competitive ratio of $1+\epsilon$.} 

While it's a significant improvement over the previous bound, we hope to do even better by exploiting the specific structure of the ski rental problem. In particular, we observe that the classification error is almost entirely due to samples close to the threshold, but for values of $y$ close to $B$, mis-classifying $z$ does not cost us significantly in the ski rental objective. This allows us to create an artificial margin around the classification boundary and discard all samples that appear in this margin. Using this improvement, we can improve the sample complexity of the training set to remove the dependence on $d$ entirely (although at a slightly worse dependence on $\eps$). 

We also consider a noisy model where the labels in the training set are noisy. By this, we mean that labels for a certain fraction of the input distribution are flipped adverserially. We design a noise tolerant algorithm for the learning-to-rent problem with a competitive ratio of $1+3\sqrt{p}$, where $p$ is the mis-classification error of a noise tolerant binary classifier. We complement this bound by showing that for a noise level of $\eta$, the best competitive ratio achievable is $1+\frac{\sqrt{\eta}}{2}$. 

Next, we consider robustness of our algorithms, i.e., their performance under no assumptions on the input. An important distinction between the recent line of research on online algorithms with predictions and previous ``beyond worst case'' approaches to competitive analysis is that the recent work simultaneously provides worst case guarantees while also improving the bounds if the additional assumptions on the input hold. Therefore, it is crucial that our algorithms are also robust in this sense. Indeed, we show that in order to obtain a competitive ratio of $(1+\eps)$ in the optimistic scenario, none of our algorithms has competitive ratios any worse than $1 + \frac{1}{\eps}$ in the adversarial setting.

% \alert{Say some more about the results in the noisy model. Then, talk about the results if the predictions are entirely correct.}

Finally, we perform numerical simulations to evaluate our learning-to-rent policies. We consider three different regimes, corresponding to small ($d=2$), moderate ($d = 100$), and large ($d = 5000$) number of feature dimensions. Recall that our margin-based technique outperforms the black box learning approach for a large number of feature dimensions. This is indeed the case in our experiments: while the two approaches are comparable for $d=2$ and exhibit relatively mild differences for $d=100$, the margin-based approach is decidedly superior for $d=5000$. In principle, this shows that in large instances, there is considerable benefit to customizing \ml predictions to make them conducive to the objectives of the online algorithm. In fact, we also show experimentally that although margin-based predictions achieve a smaller competitive ratio, their corresponding mis-classification error is rather large. This provides further evidence that a black box learning approach that simply tries to minimize classification error is not sufficient for generating good predictions for online algorithms. In addition, we also empirically evaluate the performance of our noise-tolerant algorithm and map the competitive ratio as a function of the mis-classification error.
% These empirical results provide further justification that incorporating the optimization objective in the learning algorithm leads to significant improvements. 

\smallskip\noindent{\bf Related Work.}
A robust literature is beginning to emerge in incorporating \ml predictions in online algorithms. While the list of papers in this domain continues to grow by the day, some of the representative problems that this theme has been applied to include: auction pricing~\cite{MedinaV17}, rent or buy~\cite{purohit2018improving,GollapudiP19}, caching~\cite{lykouris2018competitive,Rohatgi20,JiangPS20}, scheduling~\cite{purohit2018improving,LattanziLMV20,Mitzenmacher20}, frequency estimation~\cite{hsu2018learningbased}, Bloom filters~\cite{mitzenmacher2018model}, etc. As described earlier, these results consider \ml as a black box and re-design the online algorithm, whereas we take the complementary approach of re-designing the learning algorithm to suit the optimization task.

% There is some existing literature on “optimization from samples” (OPS) which focuses on using the given sampled values of a function (under some distribution) to optimize the function under some constraint (see \cite{}). The main difference with our work is that while OPS focuses on optimizing an unknown, learned function over a known input domain, our goal is to optimize a known function (the competitive ratio of the algorithm) for an unknown,learned input.

Our main idea is to modify the loss function in the learning algorithm to incorporate the optimization objective. There has been previous research in a similar spirit, where the loss function in learning is adapted to suit specific purposes, albeit different ones from our work. For instance, \cite{huang2019addressing} give an ``Adaptive Loss Alignment'' scheme to meta-learn the loss function to directly optimize the evaluation metric in the context of Reinforcement Learning.  \cite{gupta2017pac} present a framework for algorithm selection as a statistical learning problem. This framework captures, for instance, the notion of ``self-improving algorithms'', where the goal is to learn the input distribution and adaptively design an optimal policy (originally proposed by \cite{ailon2011self}). A related line of research, pioneered by \cite{ColeR14}, is that of optimizing on samples of the input rather than the entire input (see also \cite{MorgensternR16,BalkanskiRS16,BalkanskiRS17}). Yet another example of adapting the loss function in  learning is in Cost Sensitive Learning~\cite{elkan2001foundations}, where mis-classiﬁcation errors incur non-uniform penalties (see also \cite{kamalaruban2018minimax, ling2008cost}). 

\eat{
In formalizing our problem, we will adopt an approach based on the celebrated {\em probably approximately correct} (PAC) learning framework, introduced by \cite{valiant1984theory}. We will use upper and lower bounds relating the sample complexity of PAC-learning with the VC-dimension~\cite{vapnik1998statistical} of the concept class in binary classification~\cite{hanneke2016optimal, EhrenfeuchtHKV89}, and also extend to binary classification in the presence of noise (e.g.,~\cite{bylander1994learning,blum1998polynomial, herbrich2002pac, natarajan2013learning, awasthi2014power}).
}

\section{Preliminaries}
\label{sec:prelim}
For notational convenience, we consider a continuous version of the ski rental problem, where the buying cost is $\$ 1$, and the length of the ski season is denoted by $y$. 
(The assumption on the buying cost is w.l.o.g. by appropriate scaling.)
%The cost of buying is normalized to be 1 without loss of generality. 
Therefore, the optimal offline solution is to buy at the outset when $y\geq 1$ and rent throughout when $y<1$. 
We also denote the feature vector by $x\in \mathbb{R}^d$ (e.g., weather predictions, skier behavior, etc.)
%The variable $y$ depends on a parameter $x \in \mathbb{R}^d$. (this parameter $x$ is known as a feature and can denote the past values of $y$, the weather/ temperature etc). In this section, we are going to assume that 
and assume that $(x,y)$ is drawn from an unknown joint distribution $\mathbb{K}$. Given a feature vector $x$, the goal of the algorithm is to produce a threshold $\theta(x)$ such that the skier rents till time $\theta(x)$ and buys at that point if the ski season is longer. We call $\theta(x)$ the {\em wait time} of the algorithm.

If the distribution $\mathbb{K}$ were known to the algorithm, then for each input $x$, it can compute the conditional distribution $y|x$ and solve the resulting {\em stochastic} ski rental problem, i.e., where the input is drawn from a given distribution. It is well known that the optimal strategy in this case can be described by a fixed wait time that we denote $\theta^*(x)$. 
%and without loss of generality the optimal strategy is to set a threshold $\theta^*$, which denotes the amount of time that the algorithm decides to wait before buying - that is, the algorithm decides to rent for time $\theta(x)$, and buy if the season has not ended by then. In general, the optimal strategy for the algorithm can be described as a function $\theta^*(x)$, which maps each input $x$ to the optimal wait time $\theta^*(x)$.

Of course, in general, the distribution $\mathbb{K}$ is not known to the algorithm, and has to be ``learned'' from training data. 
The ``learning-to-rent'' algorithm observes $n$ training samples $(x_i,y_i) \sim \mathbb{K}$, and based on them, generates a function $\theta(x)$ that maps feature vectors $x$ to the wait time. %a real-number. The threshold $\theta(x)$ denotes the amount of time that the algorithm decides to wait before buying - that is, the algorithm decides to rent for time $\theta(x)$, and buy if the season has not ended by then. 
The (expected) competitive ratio of the algorithm is given by:%\footnote{Note that the function $g$ captures the competitive ratio used in analysis for online algorithms in general. This is the ratio of the cost of the online Algorithm to that of the best offline algorithm that knows the full future in advance}
\begin{align}
\comp(\theta, \mathbb{K}) &= \mathbb{E}_{(x,y) \sim \mathbb{K}}[g(\theta(x),y)] \label{eq:CR}\\
%\intertext{where $g(\theta(x),y)$ is given by:}
\text{where~} g(\theta(x),y) &= 
     \begin{cases}
      \frac{y}{\min \{y,1\}} & \text{when } y < \theta(x)\\
      \frac{1+\theta(x)}{\min\{y,1\}} & \text{when } y \ge \theta(x).\\
      \end{cases}
\end{align}

The goal of the learning-to-rent algorithm is to output a function $\theta(\cdot)$ that minimizes $\comp$ in Eq.~\eqref{eq:CR}. Since the ideal strategy is to output the function $\theta^*(\cdot)$, we measure the performance of the algorithm as the ratio between $\comp(\theta, \mathbb{K})$ and $\comp(\theta^*, \mathbb{K})$.

\begin{definition}\label{def: epsilon_accurate}
A learning-to-rent algorithm $A$ with threshold function $\theta(\cdot)$ is said be $(\epsilon, \delta)$-accurate with $n$ samples, if for any distribution $\mathbb{K}$, after observing $n$ samples, we have the following guarantee with probability at least $1 - \delta$:
%, the competitive ratio of the algorithm is at most $(1+\epsilon)$ the best achievable:
\begin{equation}\label{eq:epsilon-accurate}
    \comp(\theta, \mathbb{K}) \leq (1 + \epsilon)\cdot \comp(\theta^*, \mathbb{K}).
\end{equation}
%where $f^{*}_{\mathbb{K}} = \comp(\theta^*(x))$ is the optimal value and the optimal solution $\theta^*(x) = \arg\min_{\theta \in \mathbb{R}^+}  \mathbb{E}_{y|x}[g(\theta,y)]$.
If we say that an algorithm is $(1+\epsilon)$-accurate, we mean Eq.~\eqref{eq:epsilon-accurate} holds for some fixed constant $\delta$.
\end{definition}
The additional parameter $\mathbb{K}$ can be dropped when the distribution is clear from the context.

\section{A General Learning-to-Rent Algorithm}
\label{sec:lipschitz}
As described in the introduction, it is natural (and required) to assume that the joint distribution $\mathbb{K}$ on $(x,y)$ is {\bf Lipschitz} in the sense that similar feature vectors $x$ imply similar conditional distributions $y|x$. In this section, our main contribution is to design a learning-to-rent algorithm under this minimal assumption. 

First, we give the precise definition of the Lipschitz property we require. In particular, we measure distances between distributions using the {\em earth mover distance} (\emd) metric.

\eat{
\begin{definition}
Given two probability distributions, $P$ on set $X$ and $Q$ on set $Y$,
the \emd between $P$ and $Q$ is defined as: 
$$\argmin_{f} \int_{x}\int_{y}f(x,y)\norm{x-y}~dx~dy \quad \text{subject to}:$$
\begin{align*}
    \int_{x}f(x,y)~dy &\leq q(y),\\
    \int_{y}f(x,y)~dx &\leq p(x),\\
    \int_{x}\int_{y}f(x,y)~dx~dy &\leq 1.
\end{align*}
\end{definition}
}

\begin{definition}
For probability distributions $\mathbb{X}, \mathbb{Y}$ over $\mathbb{R}^d$,
$$\emd(\mathbb{X},\mathbb{Y}) = \min_{\mathbb{K}: \mathbb{K}\mid x = \mathbb{X}, \mathbb{K}\mid y = \mathbb{Y}} \left(\ex_{(x, y)\sim\mathbb{K}}[\norm{x-y}]\right).$$
\end{definition}

The joint distribution $\mathbb{K}$ above is such that its marginals with respect to $y$ and $x$ are equal to $\mathbb{X}$ and $\mathbb{Y}$ respectively. We now define the Lipschitz property using \emd as the distance measure between distributions.

\begin{definition}\label{def:lipchitz}
A joint distribution on $(x,y)\in \mathbb{R}^d\times \mathbb{R}^+$ is said to be $L$-Lipschitz iff for all $x_1,x_2 \in \mathbb{R}^{d}$, the marginal distributions $\mathbb{Y}_1 = y|x_1$, $\mathbb{Y}_2 = y|x_2$ satisfy $\emd(\mathbb{Y}_1, \mathbb{Y}_2) \leq L \cdot \norm{x_1 - x_2}_2$.
\end{definition}
Now we are ready to state our main result in this section:
\begin{theorem} \label{thm:lipschitz}
For the learning-to-rent problem, if $x\in [0,1]^{d}$, and the joint distribution $(x, y)$ is $L$-Lipschitz, then there exists an algorithm that uses $n = \left(\frac{L\sqrt{d}}{\epsilon}\right)^{O(d)}$ samples and is $(1 + \epsilon)$-accurate with high probability.\footnote{with probability exceeding $1 - \epsilon^{\Omega(d)}$}  % when $x\in [0,1]^{d}$ and the conditional distributions are lipchitz.
\footnote{The quantity $\epsilon$ is considered to be small ($\leq 0.01$) throughout the analysis}
\end{theorem}

\begin{algorithm}
\caption{Outputs $\theta_{A}$ for a given distribution on $y$}
\label{alg:zero_dim_x}
Query $\left(\frac{\delta}{\epsilon^6} \right)$ samples for some constant $\delta > 0$.\\ Initialize array $l$ of length $\frac{1}{\epsilon^2}$\\
Let $\ell[\theta] \leftarrow$ average of $g(\theta, y)$ over all samples $y$.\\
{\bf return} $\theta_{A} \leftarrow \argmin_{\theta \in [\epsilon,1/\epsilon], \theta/\epsilon\in \mathbb{N}} \ell [\theta]$.
\end{algorithm}

%\alert{KEERTI : READ TILL END OF SECTION}

Let us first consider a warm-up example where we have a fixed $x$ and only consider the conditional distribution $y|x$ (See Algorithm~\ref{alg:zero_dim_x}). In this case, it is natural to optimize $\theta$ over the empirical samples of $y$. However, if we don't put any constraint on $\theta$, the competitive ratio for a sample $y$ can be unbounded (this can happen when $\theta$ is close to 0 or very large), which might hurt generalization. We solve this problem by proving that it suffices to consider $\theta$ in the range $[\epsilon,1/\epsilon]$ in order to get an $(1+\epsilon)$-accurate solution. (See Lemma~\ref{lemma:restrict_theta}).

For this special case, we have the following result:

\begin{theorem}\label{thm:zero_dim_x}
If a learning-to-rent problem has only one possible input $x$, then there exists an algorithm requiring $O\left(\frac{\delta}{\epsilon^6}\right)$ samples that achieves $(1+\epsilon)$ accuracy with probability $\ge 1 -  O\left(\frac{e^{-\Omega\left(\frac{\delta}{\epsilon}\right)}}{\epsilon^2}\right)$.
\end{theorem}
\begin{algorithm}
\caption{Outputs $\theta_{A}(x)$ for multi-dimensional $x$}
\label{alg:multi_dim_x}
Divide the hyper-cube $[0,1]^d$ into sub-cubes of side length $\frac{\epsilon^3}{64L\sqrt{d}}$ each. The number of such cubes is $N = \left(\frac{64L\sqrt{d}}{\epsilon^3}\right)^{d}$. Index the cubes by $i$, where $1 \leq i \leq N$.\\
Query $\Pi = \left(\frac{1024L\sqrt{d}}{\epsilon^6}\right)^{2d}$ samples, and let $I_{\epsilon} = [\epsilon,1/\epsilon].$\\ Set threshold $\tau = \left(\frac{64L\sqrt{d}}{\epsilon^{8}}\right)^{d}$.\\
{\bf for} each sub-cube $C_i$:\\
\hspace*{10pt}{\bf if} the  number of samples from the sub-cube exceeds $\tau$\\
\hspace*{10pt}{\bf then} \\
\hspace*{15pt}Compute $\theta_{i}\leftarrow\argmin_{\theta \in I_{\epsilon}, \theta/\epsilon\in \mathbb{N} }\mathbb{E}_{(x,y) : x \in C_i}[g(\theta,y)]$.\\ 
\hspace*{15pt}For all $x \in C_i$: {\bf return} $\theta_{A}(x) \leftarrow \theta_{i}$.\\
\hspace*{10pt}{\bf else}\\ 
\hspace*{15pt}For all $x \in C_i$: {\bf return} $\theta_{A}(x) \leftarrow 1$.
\end{algorithm}

Let $\mathbb{K}$ be the distribution of $y$, and $\theta^*$ be the optimal threshold for this distribution and $f^*_\mathbb{K}$ is the optimal expected competitive ratio. We first show that it suffices to get a threshold that is not much larger than $\theta^*$:

\begin{lemma}\label{lemma:ep_close}
Let the length of the ski-renting season $y \sim \mathbb{K}$, then:
\begin{equation*}
\comp(\theta^* + \epsilon,\mathbb{K}) \leq (1 + \epsilon)f^{*}_{\mathbb{K}}\end{equation*} 
where $f^{*}_{\mathbb{K}} = \comp(\theta^*, \mathbb{K})$ is the optimal value and the optimal threshold $\theta^* = \arg\min_{\theta \in \mathbb{R}^+}  \comp(\theta, \mathbb{K})$.
\end{lemma}

\begin{proof}
We compare the competitive ratio at different values of $y$. Recall that :

\begin{equation*}
    g(\theta, y) = 
    \begin{cases*}
      \frac{(1+\theta)}{\min \{1, y\} } & if $y \geq \theta$ \\
      \frac{y}{\min \{1, y\}}        & otherwise
    \end{cases*}
\end{equation*}

When $y\leq \theta^{*}$ then both thresholds will lead to the same cost and $g(., y)$ remains unchanged. 

For $\theta^{*} + \epsilon > y > \theta^{*}$ we have 
$$\frac{g(\theta^{*} + \epsilon,y)}{g(\theta^*,y)} = \frac{y}{1 + \theta^{*}} \leq 1. $$

Finally, for $y > \theta^{*} + \epsilon$, we have $$\frac{g(\theta^{*}+\epsilon,y)}{g(\theta^*,y)} = \frac{1+\theta^{*}+\epsilon}{1 + \theta^{*}} \leq (1+\epsilon). $$

Since the ratio is bounded above by $1+\epsilon$ for all $y$, after taking the expectation we have $$\mathbb{E}_{y \sim \mathbb{K}}[g(\theta^{*} + \epsilon,y)] \leq (1+\epsilon)\cdot\mathbb{E}_{y \sim \mathbb{K}}[g(\theta^{*},y)].$$
\end{proof}

The next lemma shows that without loss of generality we only need to consider thresholds in the range $[\epsilon, 1/\epsilon]$:

\begin{lemma}\label{lemma:restrict_theta}
Let $f^{\epsilon}_{\mathbb{K}} = \min_{\theta \in \left[\epsilon,\frac{1}{\epsilon}\right]}\comp(\theta, \mathbb{K})$ then:
\begin{equation*}
f^{\epsilon}_{\mathbb{K}} \leq f^{*}_{\mathbb{K}}(1+\epsilon),
\end{equation*}
where $f^{*}_{\mathbb{K}} = \comp(\theta^*, \mathbb{K})$ is the optimal value, the optimal threshold being
$\theta^{*} = \arg\min_{\theta \in \mathbb{R}^+}  \comp(\theta, \mathbb{K})$.
\end{lemma}

\begin{proof}
Let $\theta^{\epsilon}= \argmin_{\theta \in \left[\epsilon,\frac{1}{\epsilon}\right]}$ be the optimal threshold within the range $[\epsilon,1/\epsilon]$. We consider different cases for the optimal threshold (without constraints) $\theta^*$.

\textbf{Case I}: 
When $\theta^* \in \left[\epsilon,\frac{1}{\epsilon}\right]$ then clearly we have $\theta^* = \theta^{\epsilon}$.

\textbf{Case II} : $\theta^* < \epsilon$, in this case we show that choosing $\theta = \theta^*+\epsilon$ is good enough: 
by Lemma \ref{lemma:ep_close}, we have that $\theta^* + \epsilon \in \left[ \epsilon, \frac{1}{\epsilon}\right]$ and,
$
f^{\epsilon}_{\mathbb{K}} \leq \comp(\theta^{*} + \epsilon,\mathbb{K}) \leq (1+\epsilon)f^{*}_{\mathbb{K}}$.

\textbf{Case III} : $\theta^{*} > \frac{1}{\epsilon}$, in this case we show that choosing $\theta = 1/\epsilon$ is good enough.
When $y \leq 1/\epsilon$, then $g(1/\epsilon,y) \leq g(\theta^{*},y)$.
When $y > 1/\epsilon$, then $\frac{g(1/\epsilon,y)}{g(\theta^{*},y)} \le \frac{1/\epsilon + 1}{y} \leq \frac{1/\epsilon + 1}{1/\epsilon} = 1+\epsilon$.

Hence, $f^{\epsilon}_{\mathbb{K}} \leq \mathbb{E}_{y \sim \mathbb{K}}\left[ g(1/\epsilon, y)\right] \leq (1+\epsilon)f^{*}_{\mathbb{K}}$.
\end{proof}

Next we show how to estimate the expected competitive ratio using samples from the distribution.
\begin{lemma}\label{lemma:sample}
Given a fixed $\theta \in \left[\epsilon, \frac{1}{\epsilon}\right]$, by taking $\frac{\delta}{\epsilon^4}$ samples of $y \sim \mathbb{K}$, the quantity $\mathbb{E}_{y \sim \mathbb{K}}[g(\theta,y)]$ can be estimated to a multiplicative accuracy of $\epsilon$ with probability $1 - e^{-\frac{2\delta}{\epsilon}}$.
\end{lemma}

\begin{proof}
Note that when $\theta \in \left[\epsilon, \frac{1}{\epsilon}\right]$ then 
$g(\theta, y)$ is bounded above by $\frac{1}{\epsilon} + 1$,
therefore the random variable $g(\theta,y)$ has a variance $\sigma^2$ bounded above by $\frac{1}{\epsilon^2}$. 

Let $\comp(\theta, \mathbb{K}) = \mathbb{E}_{y \sim \mathbb{K}} [g(\theta,y)]$ be the true mean of the distribution and $\widehat{\comp}(\theta, \mathbb{K})$ denotes the estimate that we have obtained by taking $\frac{\delta}{\epsilon^4}$ samples. Also, any estimate of $g(\theta, y)$ is from a distribution whose mean is $ \comp(\theta, \mathbb{K})$ and is bounded inside the range $[1,1 + \frac{1}{\epsilon}]$.
Therefore, taking $\frac{\delta}{\epsilon^4}$ samples and by Hoeffding's Inequality \cite{Hoeffding}, we claim that :
\begin{equation*}
\mathbb{P}\left[\widehat{\comp}(\theta, \mathbb{K}) - \comp(\theta, \mathbb{K}) > t \right] \leq exp\left(-\frac{2\delta t}{\epsilon^2}\right).
\end{equation*}
Setting $t=\epsilon$ and using the fact that $\comp(\theta, \mathbb{K}) \ge 1$, we get that with probability: $1 - e^{-\frac{2\delta}{\epsilon}}$, $$\widehat{\comp}(\theta, \mathbb{K}) \leq (1 + \epsilon)\comp(\theta, \mathbb{K}).$$ 
\end{proof}

Finally, we are ready to prove Theorem~\ref{thm:zero_dim_x}:
\begin{proof}[Proof of Theorem~\ref{thm:zero_dim_x}]
The algorithm simply involves dividing the segment $\left[ \epsilon, \frac{1}{\epsilon} \right]$ into small intervals of $\epsilon$ width. This would give us at most $1/\epsilon^2$ intervals.(refer to Algorithm \ref{alg:zero_dim_x}) For each interval $[\theta_0 -\epsilon,\theta_0]$ we use the $\frac{\delta}{\epsilon^6}$ samples at $\theta = \theta_0$ to calculate $\widehat{\comp}(\theta_0, \mathbb{K})$. We output the $\theta_0$ that has the minimum $\widehat{\comp}(\theta_0, \mathbb{K})$ over all such intervals.

By Lemma \ref{lemma:sample} we know that our estimate is within a $(1+\epsilon)$ multiplicative factor of the true $\comp(\theta_0, \mathbb{K})$ with probability: $1 - e^{-\frac{2\delta}{\epsilon}}$. Since there are at most $\frac{1}{\epsilon^2}$ such $\theta_0$: by a simple union bound, we claim that all our estimates on the competitive ratio are $(1+\epsilon)$ multiplicative factor of the true $\comp(\theta_0, \mathbb{K})$ with probability : $1 - \left(\frac{e^{-\frac{2\delta}{\epsilon}}}{\epsilon^2}\right)$. Also lemma \ref{lemma:ep_close} tells us that $\comp(\theta_0, \mathbb{K})$ is within a $(1+\epsilon)$ factor of $\comp(\theta, \mathbb{K})$ for all $\theta \in [\theta_0-\epsilon,\theta_0]$. Therefore, by taking the minimum over all $\theta_0$ : we are within a $(1+2\epsilon + \epsilon^2)$ factor of $\min_{ \theta \in \left[\epsilon, \frac{1}{\epsilon} \right]} \comp(\theta, \mathbb{K})$. Finally, we invoke lemma \ref{lemma:restrict_theta} to claim that our value is within a $(1+4\epsilon)$ (for $\epsilon<0.4$) multiplicative factor of $f^{*}$. Repeating the above analysis with $\epsilon' = \frac{\epsilon}{4}$, we achieve $(1+\epsilon')$ accuracy using $\frac{256 \delta}{\epsilon'^4}$ samples with probability: $1 - 16\left(\frac{e^{-8\delta/\epsilon'}}{\epsilon'^2}\right)$.
\end{proof}

To go from a single $x$ to the whole distribution, the main idea is to discretize the domain of $x$ using an $\epsilon$-net for small enough $\epsilon$.\footnote{The $\epsilon$ in the $\epsilon$-net is not the same as the accuracy parameter $\epsilon$. We are overloading $\epsilon$ in this description because the reader may be familiar with the term $\epsilon$-net; in the formal algorithm  (Algorithm~\ref{alg:multi_dim_x}), we avoid this overloading.} For each cell in the $\epsilon$-net, we show that if there are enough samples in the training set from that cell, then we can estimate the conditional probability $y|x$ to a sufficient degree of accuracy for the optimization loss to be bounded by $1+\epsilon$. On the other hand, if there are too few samples, then the probability density in the cell is small enough that it suffices to use a worst case online algorithm for all test data in the cell. (The formal algorithm is given in Algorithm~\ref{alg:multi_dim_x}.)

% We refer the reader to the full version of the paper for a formal analysis of this algorithm.
% , which establishes Theorem~\ref{thm:lipschitz}, to the appendix.

\begin{lemma}\label{closeness_in_distribution}
Given two distributions $\mathbb{D}_1,\mathbb{D}_2$ such that $EMD(\mathbb{D}_1,\mathbb{D}_2) \leq \Delta$, then:
$$\mathbb{E}_{y \sim \mathbb{D}_1}[g(\theta + \epsilon,y)] \leq (1+\epsilon)\left( 1 + \frac{\Delta}{\epsilon^2} \right) \mathbb{E}_{y \sim \mathbb{D}_2}[g(\theta ,y)], \text{ for any } \theta \in \mathbb{R}^{+}.$$
\end{lemma}

\begin{proof}
Let $p_i(y_0)$ be the probability that $y=y_0$ for distribution $\mathbb{D}_i$.
For $y \leq \theta + \epsilon$ : $\frac{g(\theta + \epsilon,y)}{g(\theta,y)} \leq 1$. 
Also, when $y> \theta+\epsilon$ then, $\frac{g(\theta + \epsilon,y)}{g(\theta,y)} = \frac{1+\theta+\epsilon}{1+\theta} \leq (1+\epsilon)$. 

Let us begin at distribution $\mathbb{D}_2$, and there be an adversary who wants to increase the expectation $\mathbb{E}_{y}[g(\theta + \epsilon, y)]$ by shifting some probability mass and thereby changing the distribution. However the adversary cannot change the distribution drastically (which is where the EMD comes into play), the total earth mover distance between the new and old distribution can be at most $\Delta$.
\begin{figure}[htb]
    \centering
    \includegraphics[width = 0.9\textwidth]{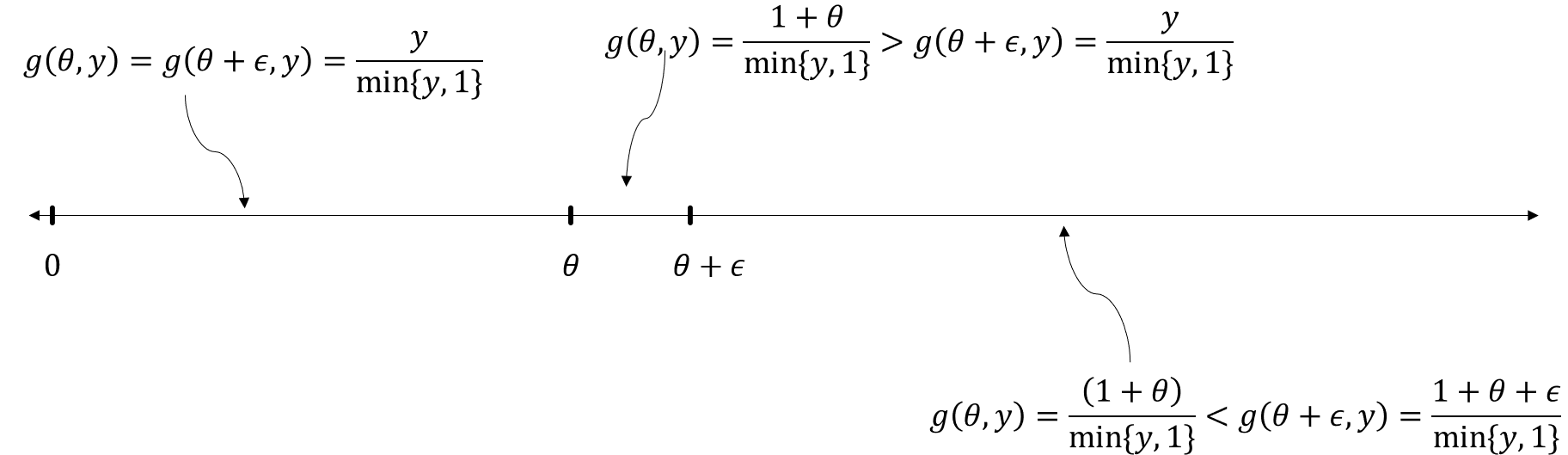}
    \caption{Value of $g(\theta, y)$ in different regimes of $y$}
    \label{fig:comp_ratio}
\end{figure}

The figure ~\ref{fig:comp_ratio} shows the different values of $g(\theta, y)$ and $g(\theta + \epsilon, y) $ in the regions where $y\leq\theta$, $y\in (\theta, \theta+\epsilon]$ and $y>\epsilon+\theta$. Note that the difference $g(\theta + \epsilon, y) - g(\theta, y)$ is greatest when $y > \epsilon + \theta$.
Shifting any probability mass within the regions $y<\theta$ or $y>\theta+\epsilon$ does not increase the quantity $\frac{g(\theta+\epsilon,y)}{g(\theta,y)}$.  
If we shift some probability mass from $y_1 \in [\theta,\theta + \epsilon]$ to $y_2 > \theta+\epsilon$, the increase in $\frac{g(\theta + \epsilon,y_2)}{g(\theta,y_1)}$ is upper bounded by $1+\epsilon$.
Note that we can shift as much mass as we want from $y_1 = \theta + \epsilon - \tau$ to $y_2 = \theta+\epsilon+\tau$ for $\tau \rightarrow 0^{+}$.

The maximum change occurs when we move from $y_1<\theta$ to $y_2>\theta+\epsilon$, then $\frac{g(\theta+\epsilon,y_2)}{g(\theta,y_1)} = \left(\frac{\min\{1,y_1\}}{y_1}\cdot\frac{(1+\theta+\epsilon)}{\min\{1,y_2\}}\right) \leq \left(\frac{(1+\theta+\epsilon)}{\min\{1,y_2\}}\right)$. However, the maximum probability mass that can be moved is upper bounded by $\frac{\Delta}{y_2 - y_1}$ (Since we know that $\mathbb{D}_1$ and $\mathbb{D}_2$ differ by $\Delta$).

Thus the upper bound we obtain is,
\begin{align*}
\frac{\mathbb{E}_{y \sim \mathbb{D}_1}[g(\theta + \epsilon,y)]}{\mathbb{E}_{y \sim \mathbb{D}_2}[g(\theta,y)]} &\leq (1+\epsilon) +  \max_{y_1\in[0,\theta) ,y_2 >\theta+\epsilon }\left(\frac{(1+\theta+\epsilon)}{\min\{1,y_2\}} \times \frac{\Delta}{y_2 -  y_1} \right)\\
&= (1+\epsilon) + \left(\frac{1+\theta+\epsilon}{\theta + \epsilon} \times \frac{\Delta}{\epsilon}\right)\\
&\leq  (1+\epsilon) + (1+\epsilon)\frac{\Delta}{\epsilon^2}\\
&=(1+\epsilon)\left(1+\frac{\Delta}{\epsilon^2} \right)
\end{align*}
\end{proof}

As a corollary, by linearity of expectation we know if many distributions are close, then their optimal solutions are also close:

%The following corollary follows from linearity of expectation.
\begin{corollary}\label{cor:relate_distributions}
Let $\mathbb{K}_1,\mathbb{K}_2$ be two joint distributions on $(X,Y)$ such that they have the same support $S$ on $X$. If $\forall x_i,x_j\in S$, $\mathbb{D}_i = Y\mid (X=x_i),\mathbb{D}_j = Y\mid (X=x_j)$ satisfies $EMD(\mathbb{D}_i,\mathbb{D}_j)\leq \Delta$ then:
\begin{equation*}
\mathbb{E}_{(x,y) \sim \mathbb{K}_1}\left[g(\theta + \epsilon,y)\right] \leq (1+\epsilon)\left (1+\frac{\Delta}{\epsilon^2} \right)\mathbb{E}_{(x,y) \sim \mathbb{K}_2}\left[g(\theta,y)\right]    
\end{equation*}
And,
\begin{equation*}
\min_{\theta} \mathbb{E}_{(x,y) \sim \mathbb{K}_1}\left[g(\theta,y)\right] \leq (1+\epsilon)\left (1+\frac{\Delta}{\epsilon^2} \right) \min_{\theta} \mathbb{E}_{(x,y) \sim \mathbb{K}_2}\left[g(\theta,y)\right]    
\end{equation*}

\end{corollary}

We now give the proof of Theorem~\ref{thm:lipschitz} to show the sample complexity to obtain a $1+\epsilon$ learning-to-rent algorithm:

\begin{proof}
Let us focus on a certain sub-cube $C_i$, we will break them into two cases: one where the sub-cube gets enough samples and one where the sub-cube does not get enough samples.

\smallskip\noindent
\textbf{CASE I}:
Let's say that we met the threshold and got over $\frac{1}{\epsilon^{8d}}$ samples in $C_i$. Let $\mathbb{K}_i$ be the true conditional distribution of $Y$ when $X=x$ lies in $C_i$. Clearly, when we are sampling $Y$ where $X$ lies inside $C_i$ our estimate might be a from a different distribution $\hat{\mathbb{K}}_i$.

But both these distributions are from a linear combinations of conditional distributions $Y\mid (X=x)$ over $x \in C_i$. Using algorithm \ref{alg:zero_dim_x} we get a $\theta_i$ for $C_i$ and using the result from Theorem \ref{thm:zero_dim_x} (with $\delta = \frac{1}{\epsilon^{8d-6}}$), and union bound over all the cubes, we can claim that with a very high probability: $\ge 1 -O\left( \epsilon^{\Omega(d)}\right)$, it satisfies:
\begin{align}
\forall i  \,\,  \mathbb{E}_{y\sim \hat{\mathbb{K}}_i}[g(\theta_i,y)] &\leq \left(1+\frac{\epsilon}{3}\right) \cdot \min_{\theta}\mathbb{E}_{y\sim \hat{\mathbb{K}}_i}[g(\theta,y)] \\
\intertext{Using Corollary \ref{cor:relate_distributions} we have the following}
\min_{\theta}\mathbb{E}_{y\sim \mathbb{K}_i}[g(\theta,y)] &\leq \left(1+\frac{\epsilon}{4}\right)\left( 1 + \frac{16\Delta}{\epsilon^2} \right) \min_{\theta}\mathbb{E}_{y\sim \hat{\mathbb{K}_i}}[g(\theta,y)]\\
\intertext{Since for any $x,y \in C$, we have $\norm{x-y}_2 \leq \frac{\epsilon^3}{64L}$, therefore using the Lipchitz assumption, we have $\Delta \leq \frac{\epsilon^3}{64}$. Hence,}
\min_{\theta}\mathbb{E}_{y\sim \hat{\mathbb{K}}_i}[g(\theta,y)] &\leq \left(1+\frac{\epsilon}{3}\right) \min_{\theta}\mathbb{E}_{y \sim \mathbb{K}_i}[g(\theta,y)].
\intertext{Using Theorem \ref{thm:zero_dim_x}, and for $\epsilon<0.1$,}
\mathbb{E}_{y\sim \mathbb{{K}}_i}[g(\theta_{i},y)] &\leq \left(1+\frac{3\epsilon}{4}\right) \min_{\theta}\mathbb{E}_{y\sim \mathbb{K}_i}[g(\theta,y)].
\end{align}

\smallskip\noindent
\textbf{CASE II}: When $C_i$ does not have enough samples to meet the threshold and we set $\theta_{A}(x) = 1$ for all $x\in C_i$. In this case, we have that $g(\theta_A(x),y) = g(1,y) \leq 2$.

We will see now that the second case occurs with a very small probability. Let $P[x \in C_i]$ be denoted as $p_i$ and let $\hat{p}_i$ be our empirical estimation of $p_i$. By Hoeffding's bound,
\begin{equation*}
    \mathbb{P}[\|p_i - \hat{p}_i \| \ge t] \leq 2 e^{-2 \Pi\cdot t^2}.
\end{equation*}

where $\Pi = \left(\frac{1024L\sqrt{d}}{\epsilon^6}\right)^{2d}$ is the number of samples we took. If we set $t = \frac{\epsilon^{4d}}{(1024L\sqrt{d})^d}$, we have:
$\norm{p_i - \hat{p}_i} < \frac{\epsilon^{4d}}{(1024L\sqrt{d})^d}$, with probability : $\ge 1 - 2exp(-\frac{2}{\epsilon^{4d}})$. By carrying a simple union bound over all such $i$, we show that the above relation holds true for all $C_i$ with probability:
$$1 - N \cdot 2 e^{-\frac{2}{\epsilon^{4d}}}.$$ 
Using simple inequalities like $e^{-x} < \frac{1}{x^2}$ for $x>0$ we can show that this probability is greater than
$\alpha = 1 - O(\epsilon^{\Omega(d)})$.

% Also since $C_i$ has more than $\frac{d^{6d}}{\epsilon^{8(d+1)}}$ samples then 
% $\hat{p}_i \geq \epsilon^{4d}$, which means that
% the probability $P[x \in C_i]$ is estimated to a multiplicative accuracy of $(1 \pm 0.01)$. with a fairly high probability of $\alppha$

Let a cube be termed \textbf{good} if it has the threshold satisfied and \textbf{bad} otherwise. Also, $C(x)$ denotes the cube which contains $x$ and $n_i$ is the number of samples lying inside cube $C_i$. Let $\mathbb{V}$ denote the discrete distribution of $x$ over the cubes.
The probability $p_i = \pr_{x\sim\mathbb{V}}[x \in C_i]$ that an $x$ chosen from $\mathbb{V}$ will lie in $C_i$ is estimated as $\hat{p}_i=\frac{n_i}{\Pi}$ and as shown above, with probability $\geq 1 - \alpha: \, \, \, \norm{p_i - \hat{p}_i} < \frac{\epsilon^{4d}}{(1024L\sqrt{d})^d}$
We obtain:
\begin{align*}
\pr_{x\sim\mathbb{V}}[C(x) \text{ is good}]&= \sum_{C_i \text{ is good}}p_i\\
&\geq \sum_{C_i \,is \,good}\frac{n_i}{\Pi} - \sum_{C_i \text{ is good}}(\|p_i -\hat{p}_i\|)\\ 
&\geq \left(\frac{\sum_{\text{all cubes } C_i}n_i - \sum_{C_i \text{ is bad}}n_i}{\Pi} \right)\\
&- \qquad \frac{\epsilon^{4d}}{(1024L\sqrt{d})^d}\times N\\
&\geq 1 -  \left(\frac{\sum_{C_i \text{ is bad}}n_i}{\Pi} \right) - \left(\frac{\epsilon}{16}\right)^{d}\\
&\geq 1 - \left(\frac{N \times \tau}{\Pi} \right) - \left(\frac{\epsilon}{16}\right)^{d}.\\
&\geq 1 - \- \left(\frac{\epsilon}{16}\right)^{d}  - \left(\frac{\epsilon}{16}\right)^{d}.
\intertext{Thus,}
\pr_{x\sim\mathbb{V}}[C(x) \text{ is bad}] \leq 2\left(\frac{\epsilon}{16}\right)^{d} \leq \frac{\epsilon}{8}.
\end{align*}
Therefore, if $\theta_{A}(x)$ is the algorithm's output and $\theta^{*}(x)$ is the optimal threshold, then we get:
\begin{align*}
\mathbb{E}_{(x,y)\sim\mathbb{K}}[g(\theta_A(x),y)]
&= \left(1+\frac{3\epsilon}{4} \right)\sum_{C_i \text{ is good}}(\min_{\theta}\mathbb{E}_{x,y \sim \mathbb{K}_i}[g(\theta,y)]\pr_{x\sim\mathbb{V}}[x \in C_i])\\
&+ \qquad 2\sum_{C_i \text{ is bad}}\pr_{x\sim\mathbb{V}}[x \in C_i])\\
&\leq \left(1+\frac{3\epsilon}{4} \right)\mathbb{E}_{(x,y)\sim\mathbb{K}}[g(\theta^{*}(x),y)] + 2\pr_{x\sim\mathbb{V}}[C(x) \text{ is bad}]\\
&\leq \left(1+\frac{3\epsilon}{4} \right)\mathbb{E}_{(x,y)\sim\mathbb{K}}[g(\theta^{*}(x),y)] + \frac{\epsilon}{4}.
\intertext{Since $\mathbb{E}_{(x,y)\sim\mathbb{K}}[g(\theta^{*}(x),y)] \geq 1$, we have}
\mathbb{E}_{(x,y)\sim\mathbb{K}}[g(\theta_A(x),y)]
&\leq \left(1+\frac{3\epsilon}{4} \right)\mathbb{E}_{(x,y)\sim\mathbb{K}}[g(\theta^{*}(x),y)] + \frac{\epsilon}{4}\cdot\mathbb{E}_{(x,y)\sim\mathbb{K}}[g(\theta^{*}(x),y)]\\
&=\left(1+\epsilon \right)\cdot \mathbb{E}_{(x,y)\sim\mathbb{K}}[g(\theta^{*}(x),y)].
\end{align*}
\end{proof}

The main shortcoming of Theorem~\ref{thm:lipschitz} is that there is an exponential dependence of the sample complexity on the number of feature dimensions $d$. Unfortunately, this dependence is necessary, as shown by the next theorem:
% whose proof also appears in the appendix.
\begin{theorem} \label{thm:lipschitz_lowerbound}
For any learning-to-rent algorithm, there exists a family of $1$-Lipschitz joint distributions $(x,y)$ where $x\in [0,1]^d$ such that the algorithm must query  $\frac{1}{\epsilon^{\Omega(d)}}$ samples in order to be $(1+\epsilon)$-accurate, for small enough $\epsilon > 0$.
%Any algorithm that is $(1+O(\epsilon))$ accurate must query $\Omega(\frac{1}{\epsilon^{3d}})$ samples when the conditional distributions are $1-lipchitz$ 
\end{theorem}

\begin{figure}[htb]
    \centering
    \includegraphics[width = 0.5\textwidth]{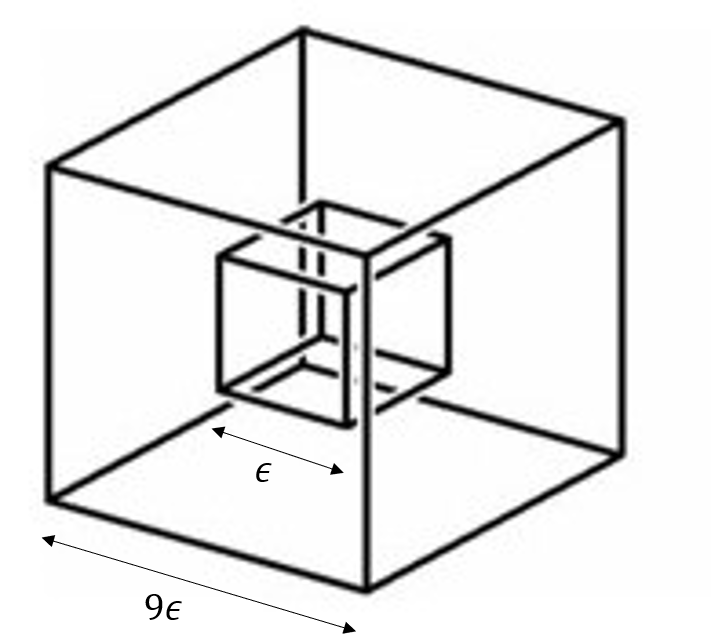}
    \caption{A sub-Hypercube with the core inside}
    \label{fig:core}
\end{figure}

In the construction, we first divide up the feature space $[0,1]^{d}$, which is in the form of a hypercube, into smaller hypercubes of side length $9\epsilon$. Note that there are $\frac{1}{(9 \epsilon)^d}$ such sub-hypercubes (see fig~\ref{fig:core}). Next, we define the {\em core} of each sub-hypercube as a hypercube of side length $\epsilon$ at the center of the sub-hypercube. In other words, the core excludes a boundary of width $4\epsilon$ in all dimensions. To reduce the effect of the $1$-Lipschitz property, we next make two design choices. First, we define $x$ as being uniformly distributed over the cores of all the sub-hypercubes, and the boundary regions have a probability density of $0$. Second, the conditional distribution $y|x$ is deterministic and invariant in any core, with value $y = 1-4\epsilon$ or $y = 1+4\epsilon$ with probability $1/2$ each. 

We now prove two key properties of this family of distributions $(x, y)$. The first lemma shows that we have effectively eliminated the information leakage caused by the $1$-Lipschitz property.

\begin{lemma}
\label{lem:lipschitz-lb}
    If an algorithm does not query any sample from a core, then it does not have any information about the conditional distribution $y|x$ in that core.
\end{lemma}
\begin{proof}
    Note that if $x_1, x_2\in \mathbb{R}^d$ are in different cores, then $\norm{x_1-x_2} \geq 8\epsilon$. This implies that even with the $1$-Lipschitz property, the \emd between the conditional distributions $y|x_1$ and $y|x_2$ can be $8\epsilon$. Since the two deterministic distributions of $y|x$ used in the construction have this \emd between them, the lemma follows.
\end{proof}

The next lemma establishes that an algorithm that does not have any information about a conditional distribution $y|x$ in any core essentially cannot do better than random guessing.

\begin{lemma}
\label{lem:cr-lb}
    If an algorithm does not query any sample from a core, then its expected competitive ratio on the conditional distribution $y|x$ in that core is at least $1 + 2\epsilon$.
\end{lemma}
\begin{proof}
    If a rent-or-buy algorithm is specified only two possible inputs where $y = 1-4\epsilon$ or $y = 1+4\epsilon$ (for small enough $\epsilon > 0$), it has two possible strategies that dominate all others: buy at time $0$ or rent throughout. The first strategy achieves a competitive ratio of $\frac{1}{1-4\epsilon} > 1+ 4\epsilon$ for $y = 1-4\epsilon$ and $1$ or $y = 1+4\epsilon$, whereas the second strategy achieves a competitive ratio of $1$ for $y = 1-4\epsilon$ and $1+4\epsilon$ or $y = 1+4\epsilon$. Since the two conditional distributions are equally likely in a core for the family of joint distributions constructed above, the lemma follows.
\end{proof}

We are now ready to prove Theorem~\ref{thm:lipschitz_lowerbound} 
% which we restate for completeness:
% \begin{theorem*}
% For any learning-to-rent algorithm, there exists a family of $1$-Lipschitz joint distributions $(x,y)$ where $x\in [0,1]^d$ such that the algorithm must query  $\frac{1}{\epsilon^{\Omega(d)}}$ samples in order to be $(1+\epsilon)$-accurate, for small enough $\epsilon > 0$.
% %Any algorithm that is $(1+O(\epsilon))$ accurate must query $\Omega(\frac{1}{\epsilon^{3d}})$ samples when the conditional distributions are $1-lipchitz$
% \end{theorem*}

\begin{proof}[Proof of Theorem~\ref{thm:lipschitz_lowerbound}]
%Let us say that the algorithm achieves a multiplicative accuracy of $(1+k\epsilon)$. We will show that there exists a constant $C$ (dependent on $k$) such that the algorithm has to take greater than $\frac{1}{C^{3d}\epsilon^{3d}}$ samples.
Assume, if possible, that the algorithm uses $n = 1/\epsilon^{d/4}$ samples. Recall that we have $1/(9\epsilon)^d > 1/\epsilon^{d/2} = n^2$ sub-hypercubes (for small enough $\epsilon$) in the construction above. This implies that for at least $1-1/n$ fraction of the sub-hypercubes, the algorithm does not get any sample from them. By Lemmas~\ref{lem:lipschitz-lb} and ~\ref{lem:cr-lb}, the competitive ratio of the algorithm on these sub-hypercubes is no better than $1 + 2\epsilon$. Therefore, even if the algorithm achieved a competitive ratio of $1$ on the other sub-hypercubes. the overall competitive ratio is no better than $(1-1/n)\cdot (1+2\epsilon) + (1/n)\cdot 1 > 1 + \epsilon$. The theorem now follows from the observation that an optimal algorithm that knows the conditional distributions $y|x$ in all sub-hypercubes achieves a competitive ratio of $1$.
\end{proof}

\eat{
The basic idea in this theorem is that with a small number of samples, there are large parts of the feature space that the algorithm does not have any samples from. In these parts, the algorithm has no information about the conditional distribution $y|x$ and has to resort to essentially random guessing, which results in a worse competitive ratio. Concretely, if we partition the feature space into a multi-dimensional array of sub-hypercubes, then most of the hypercubes have no sample and therefore, the algorithm has no information about the value of $y$ in these hypercubes. However, in implementing this strategy, we need to be careful because of the $1$-Lipschitz condition. In particular, the algorithm can derive some information about a region of the feature space that it has no samples from, simply by extrapolating from nearby samples. We handle this difficulty by focusing the probability mass of a hypercube in a smaller {\em core} that is sufficiently separated from the boundary ensuring that the value of $y$ in the core is independent of that in a neighboring hypercube. We give details of this construction and the resulting proof of Theorem~\ref{thm:lipschitz_lowerbound} in the appendix.
}

\section{A PAC Learning Approach to the Learning-to-Rent Problem}
\label{sec:pac}
In the previous section, we saw that without making further assumptions, the number of samples required by a learning-to-rent algorithm will be exponential in the dimension of the feature space. %We are thus motivated to search for certain reasonable assumptions on our problem to yield better algorithms.
To avoid this, we try to identify reasonable assumptions that allow the learning-to-rent algorithm to be more efficient.

We follow the traditional framework of PAC learning. Recall that in PAC learning, the true function mapping features to labels is restricted to a given {\em concept class} $\mathcal{C}$:

\begin{definition}
Consider a set $X \in \mathbb{R}^{d}$ and a concept class $\mathcal{C}$ of Boolean functions $X \rightarrow \{0, 1 \}$. Let $c$ be an arbitrary hypothesis in $\mathcal{C}$. Let $P$ be a PAC learning algorithm that takes as input the set $S$ comprising $m$ samples $(x_i, y_i)$ where $x_i$ is sampled from a distribution $\mathbb{D}$ on $X$ and $y_i = c(x_i)$, and outputs a hypothesis $\hat{c}$. $P$ is said to be have $\epsilon$ error with failure probability $\delta$, if with probability at least $1 - \delta$:
\begin{align*}
    \mathbb{P}_{x \sim \mathbb{D}}[\hat{c}(x) \neq c(x)] &\leq \epsilon. 
\end{align*}
\end{definition}

Standard results in learning theory show that if the function class $\mathcal{C}$ is ``simple'', the PAC-learning problem can be solved with a small number of samples. In the learning-to-rent problem, our goal is to learn the optimal policy $\theta^*(\cdot)$.

We consider the situation where the value of $y$ is deterministic given $x$. This assumption says that the features contain enough information to predict the length of the ski season. 
\begin{assumption} \label{assump:determ}
In the input distribution $(x,y)\sim \mathbb{K}$ for the learning-to-rent algorithm, the value of $y$ is a deterministic function of $x$ i.e $y = f(x)$ for some function $f$.
\end{assumption}
Note that in this case, the optimal solution is going to have competitive ratio of 1, so an $(1+\epsilon)$-accurate learning-to-rent algorithm 
must have competitive ratio $1+\epsilon$.

Because of Assumption~\ref{assump:determ}, the entire feature space can be divided into two regions: one where $y<1$ and renting is optimal, and the other where $y\ge1$ and buying at the outset is optimal. If the boundary between these two regions is PAC-learnable, we can hope to improve on the result from the previous section. This could also be seen as a weakening of Assumption~\ref{assump:determ}:

%Let $b: \mathbb{R}^{+} \mapsto \{0,1\}$ be a binary function such that $b(x) = 1$ when $y\geq 1$ and $b(x) = 0$ otherwise. 
%We will assume that $b(x)$ is in a simple function class:

%Our \textbf{first assumption} is that the binary variable $Z = B(x)$ is uniquely determined from the input parameter $x$, that is for a given $x$, $Z$ is either $0$ or $1$ but its fixed.

\begin{assumption}\label{assump:class}
In the input distribution $(x,y)\sim \mathbb{K}$ for the learning-to-rent algorithm where $X$ is the domain for $x$, there exists a hypothesis $c: X \mapsto \{0, 1 \}$ lying in a concept class $\mathcal{C}$ such that $c$ separates the regions $y\geq 1$ and $y < 1$. For notational purposes, we say $c(x) = 1$ when $y\geq 1$ and $c(x) = 0$ when $y < 1$.
\end{assumption}

\smallskip\noindent{\bf PAC-learning as a black box.}
We first show that in this setting, one can use the PAC-learning algorithm as a black-box. In other words, if we can PAC-learn the concept class $\mathcal{C}$ accurately, then we can get a competitive algorithm for the ski-rental problem.
The precise algorithm is given in Algorithm~\ref{algo:naive-pac}. Note that we only use Assumption~\ref{assump:class} here.%, that only requires the binary variable $(y>1)$ to be deterministic in $x$. 

\begin{algorithm}
\caption{Black box learning-to-rent algorithm}
\label{algo:naive-pac}
Set $\tau = \sqrt{\epsilon}$\\

{\bf Learning:} Query $n$ samples. Train a PAC-learner.\\
%\If{sample $y$ belongs to $[1- \gamma, 1 + \gamma]$}
%{Discard}
%\Else{Feed sample to PAC learner with $\alpha$-margin}

{\bf For test input $x$:}\\
{\bf if} PAC-learner predicts $y\geq 1$\\
{\bf then} $\theta(x) = \tau$\\
{\bf else} $\theta(x) = 1$.
\end{algorithm}

The next theorem relates the competitive ratio achieved by Algorithm~\ref{algo:naive-pac} with the accuracy of the black-box PAC learner.
This implies an upper bound on the sample complexity of learning-to-rent, using the sample complexity bounds for PAC learners.
% The proof is deferred to the appendix.

% \alert{KEERTI : READ TILL END OF SECTION }

\begin{theorem}\label{thm:pac-blackbox}
Given an algorithm that PAC-learns the concept class $\mathcal{C}$ with error $\epsilon$ and failure probability $\delta$, 
%under Assumptions~\ref{assump:determ} and \ref{assump:class}, 
there exists a learning-to-rent algorithm that has a competitive ratio of $(1+2\sqrt{\epsilon})$ with probability $1 - \delta$.
\end{theorem}
% \noindent{\bf Remark:}
% The above theorem can be refined for asymmetrical errors (where the classification errors on the two sides are different) showing that the algorithm is more sensitive to errors of one type than the other. 
% Further details on this appear in the appendix.
\begin{proof}%[Proof of Theorem~\ref{thm:pac-blackbox}]
The algorithm first uses PAC-learning as a black box to learn a hypothesis $\hat{c}$. We then set $\theta(x) = 1$ when $\hat{c}(x) = 0$ and setting $\theta(x)=\tau$ (for some small $\tau$ that we fix later) when $\hat{c}(x) = 1$.

If $\mathbb{D}$ denotes the distribution of input parameter $x$ then we know that,
\begin{align}
    \pr_{x\sim D}[c(x) \neq \hat{c}(x)] &\leq \epsilon.
\end{align}
Obviously, when $\hat{c}(x) = c(x)=1$, then our worst-case competitive ratio is $1+\tau$. When $\hat{c}(x) = c(x)=0$, then our competitive ratio is $1$.
Also with probability $\epsilon$, $c(x) \not= \hat{c}(x)$ and the worst case competitive ratio is $\max(2, 1 + 1/\tau)$.

If we use $\tau = \sqrt{\epsilon}$, we see that the competitive ratio $CR$ is bounded above as:
\begin{align*}
    \comp(\theta, \mathbb{K}) &\leq \left(1 + \frac{1}{\tau}\right)\cdot\epsilon + (1-\epsilon)\cdot(1 + \tau)\\
    &= 1 + \frac{\epsilon}{\tau} + \tau\cdot(1-\epsilon)
    \leq 1 + 2\sqrt{\epsilon}.
\end{align*}
Hence, with probability $1 - \delta$, we achieve a competitive ratio of $(1 + 2\sqrt{\epsilon})$.
The robustness bounds follows immediately from Lemma~\ref{lemma:robust} by noting that $\theta \ge \sqrt{\epsilon}$ for all inputs.
\end{proof}

The above result can be refined for asymmetrical errors (where the classification errors on the two sides are different) showing that the algorithm is more sensitive to errors of one type than the other. 
Let us first define a PAC learner with asymmetrical errors as follows:

\begin{definition}
Given a set $X \in \mathbb{R}^{d}$ and a concept class $C$ of Boolean functions $X \rightarrow \{0, 1 \}$. Let there be an arbitrary hypothesis $c \in C$. Let $P$ be a PAC learning algorithm that takes as input the set $S$ comprising of $m$ samples $(x_i, y_i)$ where $x_i$ is sampled from a distribution $\mathbb{D}$ on $X$ and $y_i = c(x_i)$, and outputs a hypothesis $\hat{c}$. $P$ is said to be have an $(\alpha,\beta)$ error with failure probability $\delta$, if with probability at least $1 - \delta$ on the sampling of set $S$.
\begin{align*}
    \mathbb{P}_{x \sim \mathbb{D}}[\hat{c}(x)= 0, c(x)=1] &\leq \alpha\\
    \mathbb{P}_{x \sim \mathbb{D}}[\hat{c}(x)= 1, c(x)=0] &\leq \beta
\end{align*}
\end{definition}

We can now show a better bound on the competitive ratio given access an asymmetrical learner.
\begin{theorem}
Given an algorithm that PAC learns the concept class $C$ with asymmetrical errors $(\alpha,\beta)$ and failure probability $\delta$, there exists an algorithm that has a competitive of $(1+3\epsilon)$ with probability $1 - \delta$, where $\epsilon = \max(\alpha, \sqrt{\beta})$ 
\end{theorem}

\begin{proof}
Again we use PAC-learning as a black box to learn a hypothesis $\hat{c}$. We then set $\theta(x) = 1$ when $\hat{c}(x) = 0$ and setting $\theta(x)=\tau$ (for some small $\tau$ that will be decided later) when $\hat{c}(x) = 1$

Note that with probability $\alpha$, $\hat{c}(x)=0$ and $c(x)=1$, then we have competitive ratio being capped at $2$.
And with probability $\beta$, $\hat{c}(x)=1$ and $c(x)=0$, and our competitive ratio in this case is $\frac{1+\tau}{\tau}$.
The rest of the cases, we have the competitive ratio capped at $1 + \tau$.

The expected $\comp$ is therefore,
\begin{align*}
    \comp &\leq \beta(1 + \frac{1}{\tau}) + 2\alpha +(1-\alpha - \beta)(1 + \tau)\\
    &=1 + \alpha(1-\tau) + \tau + \frac{\beta}{\tau}\\
    &\leq 1 + \alpha + \tau + \frac{\beta}{\tau}
\end{align*}
The $\comp$ is minimized at $\tau = \epsilon = \max ( \alpha, \sqrt{\beta})$ and its value is $1 + 3\epsilon$. 
\end{proof}

%\smallskip\noindent
%{\bf Relationship between PAC-learning and learning-to-rent.}
Next, we show that the relationship between PAC-learning and learning-to-rent, established in one direction in Theorem~\ref{thm:pac-blackbox}, actually holds in other direction too. In other words, we can derive a PAC-learning algorithm from a learning-to-rent algorithm. This implies, for instance, that existing lower bounds for PAC-learning also apply to learning-to-rent algorithms. Therefore, in principle, the sample complexity of the algorithm in Theorem~\ref{thm:pac-blackbox} is (nearly) optimal without any further assumptions. 
% The proof of this theorem is deferred to the appendix.

\begin{theorem}
\label{thm:naive-lb}
If there exists an $(\epsilon,\delta)$-accurate learning-to-rent algorithm for a concept class $\mathcal{C}$ with $n$ samples, then there exists an $\left(4\epsilon,\delta\right)$ PAC-learning algorithm for $\mathcal{C}$ with the same number of samples.
\end{theorem}
\begin{proof}%[Proof of Theorem~\ref{thm:naive-lb}]
We will design a PAC learning algorithm (call it $P$) using the learning-to-rent algorithm (call it $A$).
Given a sample $(x_i,z_i)$ for $P$, we define sample $(x_i,y_i)$ for $A$ where $y_i = 10$ when $z_i=1$, and $y_i = 0$ or $y=\frac{1}{2}$ with probability $\frac{1}{2}$ each, when $z_i=0$.
The output for $P$ for a feature $x$ is decided as follows:
when $\theta(x) \ge \frac{1}{2}$ predict $\hat{z}=0$, otherwise, predict $\hat{z}=1$.

First, we calculate the probability $\mathbb{P}[\hat{z}=0, z=1]$.
When $P$ predicts $\hat{z}=0$, then we have $\theta(x) \ge \frac{1}{2}$. But if, $z=1$, then the optimal cost is $1$, whereas the algorithm pays at least $\frac{3}{2}$. Hence the competitive ratio is bounded below by $\frac{3}{2}$. Since the overall competitive ratio is less than $(1+\epsilon)$, we have that: 
$$(1- \mathbb{P}[\hat{z}=0, z=1]) +  (3/2)\cdot \mathbb{P}[\hat{z}=0, z=1] \leq (1+\epsilon).$$ 
Therefore,
$\mathbb{P}[\hat{z}=0, z=1] \leq 2\epsilon$.

Second, we calculate the error on the other side which is the probability $\mathbb{P}[\hat{z}=1, z=0]$.
When $P$ predicts $\hat{z}=1$, then we have $\theta(x) < \frac{1}{2}$. But if $z=0$, then $y=\frac{1}{2}$ with probability $\frac{1}{2}$, and therefore, the competitive ratio is $\ge 2$ with probability $\frac{1}{2}$. With the remaining probability of $\frac{1}{2}$, we have $y=0$, and therefore, the competitive ratio is $\ge 1$. Therefore, the overall competitive ratio when $\hat{z}=1, z=0$ is $\ge \frac{2+1}{2} = \frac{3}{2}$. As earlier, we have
$\mathbb{P}[\hat{z}=1, z=0] \leq 2\epsilon$.
$\mathbb{P}[\hat{z} \neq z] = \mathbb{P}[\hat{z}=1, z=0] + \mathbb{P}[\hat{z}=0, z=1] \leq 4\epsilon$.
\end{proof}

%\begin{corollary}
%The sample complexity in Theorem~\ref{thm:pac-blackbox} is asymptotically optimal.
%\end{corollary}

% Now we will show that we do not necessarily need to learn the hypothesis so well to get a good algorithm.

\subsection{Margin-based PAC-learning for Learning-to-Rent}
Theorem~\ref{thm:pac-blackbox} is very general in that there are many concept classes for which we have existing PAC-learning bounds. On the other hand, even for a simple linear separator, PAC-learning requires at least $\Omega(d)$ samples in $d$ dimensions, which can be costly for large $d$. However, the sample complexity can be reduced when the VC-dimension of the concept class is small:
\begin{theorem}[e.g., \cite{kearnsV}]
\label{thm:pac-vc}
    A concept class of VC-dimension $D$ is $(\epsilon, \delta)$ PAC-learnable using $n = \Theta\left(\frac{D + \log (1/\delta)}{\epsilon}\right)$ samples. For fixed $\delta$, the sample complexity of PAC-learning is $\Theta\left(\frac{D}{\epsilon}\right)$.
\end{theorem}

In particular, this result is used when the 
underlying data distribution has a {\em margin}, which is the distance of the closest point to the decision boundary:

\begin{definition}
Given a data set $D \in \mathbb{R}^{d} \times \{0, 1\}$ and a separator $c$, the margin of $D$ with respect to $c$ is defined as 
$\min_{x' \in \mathbb{R}^{d}, (x, y)\in D, c(x')\neq y} \norm{x-x'}$.
\end{definition}

The advantage of having a large margin is that it reduces the VC-dimension of the concept class. Since the precise dependence of the VC-dimension on the width of the margin (denoted $\alpha$) depends on the concept class $\cal C$, let us denote the VC-dimension by $D(\alpha)$.

Crucially, we will show that in the learning-to-rent algorithm, it is possible to {\em reduce the sample complexity even if the original data $(x,y)\sim\mathbb{K}$ does not have any margin}. The main idea is that the learning-to-rent algorithm can ignore training data in a suitably chosen margin. This is because $y\approx 1$ for points in the margin, and the competitive ratio of ski rental is close to $1$ for these points even with no additional information. Thus, although the algorithm fails to learn the label of test data near the margin reliably, this does not significantly affect the eventual competitive ratio of the learning-to-rent algorithm. 

Note that the $L$-Lipschitz property under Assumption~\ref{assump:determ} is:
\begin{assumption}\label{assump:lipchitz}
For $x_1,x_2\in X$ where $X$ is the domain of $x$, if $y_1 = f(x_1)$ and $y_2 = f(x_2)$, we have
$|y_1 - y_2 | \le L \cdot \norm{x_1-x_2}$.
%Given $x_1,x_2 \in X$, such that $\norm{x_1 - x_2} \leq \alpha$, then $|y_1 - y_2| < L\alpha$.
\end{assumption}

We now give a learning-to-rent algorithm that uses this margin-based approach (Algorithm~\ref{algo: margin-based-pac}). Recall that $\alpha$ is the width of the margin used by the algorithm; we will set the value of $\alpha$ later.

\begin{algorithm}
\caption{Margin-based learning-to-rent algorithm}
\label{algo: margin-based-pac}
Set $\gamma = L\alpha$.\\

{\bf Learning:} Query $n$ samples. Discard samples $(x_i, y_i)$ where $y_i\in [1-\gamma, 1+\gamma]$. Use the remaining samples to train a PAC-learner with margin $\alpha$.\\
%\If{sample $y$ belongs to $[1- \gamma, 1 + \gamma]$}
%{Discard}
%\Else{Feed sample to PAC learner with $\alpha$-margin}

{\bf For test input $x$:}\\
{\bf if} PAC-learner predicts $y\geq 1$\\
{\bf then} $\theta(x) = \gamma$\\
{\bf else} $\theta(x) = 1+\gamma$.
\end{algorithm}

The filtering process creates an artificial margin:
\begin{lemma}
\label{lem:margin}
In Algorithm~\ref{algo: margin-based-pac}, the samples used in the PAC learning algorithm have a margin of $\alpha$.
\end{lemma}

We now analyze the sample complexity of Algorithm~\ref{algo: margin-based-pac}.
\begin{theorem}
\label{thm:margin}
Given a concept class $\mathcal{C}$ with VC-dimension $D(\alpha)$ under margin $\alpha$, there exists a learning-to-rent algorithm that has a competitive ratio of $1 + O(L\alpha)$ for $n$ samples with constant failure probability, where $\alpha$ satisfies:
\begin{equation}
    \label{eq:alpha}
    \sqrt{\frac{D(\alpha)}{n}} = L \alpha.
\end{equation}
\end{theorem}
\begin{proof}
Let $q$ denote the probability that $(x_i, y_i)$ satisfies $1- \gamma \leq y_i \leq 1+\gamma$, i.e., is in the margin. With probability $1- q$, a test input does not lie in the margin and we have the following two scenarios:
\begin{itemize}
\item With probability $(1 - \epsilon)$, the prediction is correct and the competitive ratio is at most $(1 + \gamma)$.
\item With probability $\epsilon$, the prediction is incorrect and the competitive ratio is at most $\max \left(1 + \frac{1}{\gamma}, 2+\gamma\right)$. For small $\gamma$ (say $\gamma \leq 1/2$, which will hold for any reasonable sample size $n$), this value is $1 + \frac{1}{\gamma}$. 
\end{itemize}
With probability $q$, a test input lies in the margin and the competitive ratio is at most $\frac{1+\gamma}{1-\gamma}$. The expected competitive ratio is:
\begin{align*}
&\comp(\theta,\mathbb{K}) 
\leq (1-q)\cdot(1 - \epsilon)\cdot(1 + \gamma) +\\
& \qquad \qquad \quad + (1-q)\cdot\epsilon\cdot\left(1+\frac{1}{\gamma}\right) + q\cdot\left(\frac{1+\gamma}{1 - \gamma}\right)\\
&\leq 1 + \left[ (1-q)\cdot(1 - \epsilon)\cdot\gamma +
(1-q)\cdot\epsilon\cdot\frac{1}{\gamma} + q\cdot\frac{2\gamma}{1 - \gamma}\right]\\
&\leq 1 + 4\gamma + (1-q)\cdot\frac{\epsilon}{\gamma}  \qquad \text{for $\gamma \leq 1/2$}.
\end{align*}

Now, note that by Chernoff bounds (see, e.g., \cite{MotwaniR97}), the number of samples used for training the
classifier after filtering is $n_f \geq n(1-q)/2$ with constant probability.
\eat{
\[
    \zeta = \mathbb{P}\left[n_f < \frac{n(1-q)}{2}\right] \le  \left(\frac{e}{2}\right)^{-\frac{n(1-q)}{2}}.
\]
Therefore, we get with probability $1- \zeta$
\[
    n_f \ge \frac{n(1-q)}{2}.
\]
}
Also, by Theorem \ref{thm:pac-vc} and Lemma~\ref{lem:margin}, we predict whether $y<1$ or $y\ge 1$ with an error rate of $\epsilon = O\left(\frac{D(\alpha)}{n_f}\right)$ using $n_f$ samples with constant probability. This implies:
$$(1-q)\cdot\epsilon = O\left(\frac{D(\alpha)}{n}\right).$$
Thus, 
$\comp(\theta,\mathbb{K})\leq 1 + 4\gamma + O\left(\frac{D(\alpha)}{n\cdot\gamma}\right)$.
Optimizing for $\gamma$, we have $\gamma = \theta\left(\sqrt{\frac{D(\alpha)}{n}}\right)$.
But, we also have $\gamma = L\alpha$ in the algorithm. This implies that we choose $\alpha$ to satisfy Eq.~\eqref{eq:alpha} and obtain a competitive ratio of $1 + O(L\alpha)$.
%For the robustness bound, note that by Lemma~\ref{lemma:robust}, the minimum $\theta$ set by the algorithm is $\gamma$, which yields a robustness bound of $1+\frac{1}{\gamma} = 1+\left(\frac{1}{L\alpha}\right)$
%
\eat{

There are two cases here: Either $p<1 - 10\cdot\frac{\log n}{n}$ and therefore, with probability $\ge 1 - O(\frac{1}{n})$:
\begin{align*}
CR&\le 1 + 10\gamma + \left(1 + \frac{1}{\alpha^2}(1-p) \right)\frac{2}{n(1-p)\gamma}.\\
&= 1 + 10\gamma + \left(1 + \frac{1}{\alpha^2}\right)\frac{2}{n\gamma}.
\end{align*}
The other case is when $p\ge 1 - 10\left(\frac{\log n}{n}\right)$, since we know that $\epsilon<1$, therefore,\\
$CR \leq 1 + 10\gamma + \frac{20\log n}{n\gamma}$.
Choosing $\gamma = L\alpha = \frac{\sqrt{L}}{n^{1/4}}$, we get for either case,
$CR \leq 1 + O\left(\frac{L}{n^{1/4}}\right)$
}
\end{proof}

We now apply this theorem for the important and widely used case of linear separators. The following well-known theorem establishes the VC-dimension of linear separators with a margin.

%When the concept class $\mathcal{C}$ is the set of linear separators (hyperplanes), it is well-known that the VC-dimension of a linear separator with margin can be much lower: %following theorem is taken from Theorem 8.4 from the Vapnik's Book \cite{vapnik1998statistical}. It relates the VC-dimension of a concept class (specifically, set of hyper-planes) to the length of the margin.
\begin{theorem}[see, e.g., \cite{vapnik1998statistical}]
\label{thm:margin_vc}
For an input parameter space $X \in \mathbb{R}^{d}$ that lies inside a sphere of radius $R$, the concept class of $\alpha$-margin separating hyper-planes for $X$ has the VC dimension $D$ given by:
$$D \leq \min\left(\frac{R^2}{\alpha^2}, d \right)+1.$$
\end{theorem}
Feature vectors are typically assumed to be normalized to have constant norm, i.e., $R=O(1)$. Thus, Theorem~\ref{thm:margin} gives the sampling complexity for linear separators as follows:
\begin{corollary}
\label{cor:hyperplane}
    For the class of linear separators, there is a learning-to-rent algorithm that takes as input $n$ samples and has a competitive ratio of $1 + O\left(\frac{\sqrt{L}}{n^{1/4}}\right)$.
\end{corollary}
\eat{
\begin{proof}
    Using Theorem~\ref{thm:margin_vc}, we rewrite Eq.~\eqref{eq:alpha} as
    $\frac{1}{\alpha\sqrt{n}} = L\alpha$, which means $L\alpha = \frac{\sqrt{L}}{n^{1/4}}$.
    The corollary follows from Theorem~\ref{thm:margin}.
\end{proof}
}

For instances where a linear separator does not exist, a popular technique called {\em kernelization} (see \cite{rasmussen2003gaussian}), is to transform the data points $x$ to a different space $\phi(x)$ where they become linearly separable.  
\begin{corollary}
For a kernel function $\phi$ satisfying $\norm{\phi(x_1) - \phi(x_2)} \geq \frac{1}{\nu}\cdot\norm{x_1-x_2}$ for all $x_1, x_2$, assuming the data is linearly separable in kernel space, 
there exists a learning-to-rent algorithm that achieves a competitive ratio of $1+O\left(\frac{\sqrt{L\nu}}{n^{1/4}}\right)$ with $n$ samples, \end{corollary}

Conceptually, the corollary states that we can make use of these kernel mappings without hurting the competitive ratio bounds achieved by the algorithm. This is because the sample complexity in the margin-based algorithm (Algorithm~\ref{algo: margin-based-pac}) is independent of the number of dimensions.

\eat{
\textcolor{blue}{:might want to compress the following to definitions into one?}
\begin{definition}
For a hypothesis $c \in \mathcal{C}$, the distance of a point $x \in \mathbb{R}^{d}$ from $c$ is denoted $d(x, c)$ is defined as :
\begin{equation*}
     \min_{y \in \mathbb{R}^{d}, c(x)\neq c(y)} \left\{\norm{x-y} \right \} 
\end{equation*}
\end{definition}
Now we are ready to define a margin as follows:
\begin{definition}
Given a concept class $\mathcal{C}$ and a set of points $X=\{x_1, x_2, \ldots x_n\}$ the margin of $(X,C)$ is defined as : 
\begin{equation*}
     \max_{c \in C} \min_{x \in X}  d(x, c) 
\end{equation*}
\end{definition}
}

\section{Learning-to-rent with a Noisy Classifier}
\label{sec:pac-learning-noise}
So far, we have seen that PAC-learning a binary classifier with deterministic labels (Assumption~\ref{assump:determ}) 
is sufficient for a learning-to-rent algorithm. However, in practice, the data is often noisy,
which leads us to relax Assumption~\ref{assump:determ} in this section. 
Instead of requiring $y|x$ to be deterministic, we only insist that $y|x$ is predictable with sufficient probability.
In other words, we replace Assumption~\ref{assump:determ} with the following (weaker) assumption:

\smallskip\noindent
{\bf Assumption 1'.}
{\em In the input distribution $(x,y)\sim \mathbb{K}$, there exists a deterministic function $f$ and a parameter $p$ such that the conditional distribution of $y|x$ satisfies $y = f(x)$ with probability at least $1-p$.}%the value of $y$ is a deterministic function of $x$ ($y = f(x)$ for some function $f$)

\smallskip\noindent
This definition follows the setting of binary classification with noise first introduced by \cite{bylander1994learning}.
Indeed, the existence of noise-tolerant binary classifiers (e.g.,~\cite{blum1998polynomial,awasthi2014power, natarajan2013learning}), leads us to ask if these classifiers can be utilized to design learning-to-rent algorithms under Assumption~1'. 
We answer this question in the affirmative by designing a learning-to-rent algorithm in this noisy setting
(see Algorithm~\ref{alg:noisy}). This algorithm assumes the existence of a binary classifier than can 
tolerate a noise rate of $p$ and achieves classification error of $\epsilon$. Let $p_0 = \max(p,\epsilon)$.
If $p_0$ is large, then the noise/error rate is too high for the classifier to give reliable information 
about test data; in this case, the algorithm reverts to a worst-case (randomized) strategy. On the other
hand, if $p_0$ is small, the the algorithm uses the label output by the classifier, but with 
a minimum wait time of $\sqrt{p_0}$ on all instances to make it robust to noise 
and/or classification error.

\begin{algorithm}
\caption{Learning-to-rent with a noisy classifier}
\label{alg:noisy}
Set $p_0 = \max(p,\epsilon)$.\\

{\bf Learning:} \\
{\bf if} $p_0 \le \frac{1}{9(e-1)^2}$\\ 
{\bf then} PAC-learn the classifier on $n$ (noisy) training samples.\\

{\bf For test input} $x$:\\
{\bf if} $p_0 > \frac{1}{9(e-1)^2}$\\
{\bf then} $\pr[\theta(x) = z] = \begin{cases} \frac{e^{z}}{e-1},~\text{for}~z\in [0,1]\\ 0,~\text{for}~ z> 1.\end{cases}$\\ %and with rest of the constant mass at $x=1$}
{\bf else}\\
\hspace*{10pt}{\bf if} PAC-learner predicts $y<1$\\
\hspace*{10pt}{\bf then} $\theta(x) = 1$\\
\hspace*{10pt}{\bf else} $\theta(x) = \sqrt{p_0}$.
\end{algorithm}

The next theorem shows that this algorithm has a competitive ratio of $1+O(\sqrt{p_0})$
for small $p_0$, and does no worse than the worst case bound of $\frac{e}{e-1}$ irrespective
of the noise/error: 
% The proof of the theorem is deferred to the appendix.

% \alert{KEERTI : READ TILL END OF SECTION }

\begin{theorem}
\label{thm:pac-noise}
%Under the modified assumption 1', and \ref{assump:class}, 
If there is a PAC-learning algorithm that can tolerate noise of $p$ and achieve accuracy $\epsilon$, the above algorithm achieves a competitive ratio of $\min(1 + 3\sqrt{p_0},\frac{e}{e-1})$ where $p_0 = \max\{p,\epsilon\}$.
\end{theorem}

\begin{proof}[Proof of Theorem~\ref{thm:pac-noise}]
Note that if $p_0 > \frac{1}{9(e-1)^2}$, we choose the threshold $\theta$ according to:
$$
\Pr[\theta = z] = \begin{cases} \frac{e^{z}}{e-1},~\text{for}~z\in [0,1]\\ 0,~\text{for}~ z> 1.\end{cases}
$$
It can be verified by taking the expectation that $\mathbb{E}[Alg] = \frac{e}{e-1}\times \min\{y,1\}$ and we obtain the competitive ratio $\frac{e}{e-1}$ \cite{KMMO94}. We now assume that $p_0 < \frac{1}{9(e-1)^2}$ for the rest of the proof.

We first focus on the points where the PAC learner's prediction is correct. This is indeed true for $1-\epsilon$ fraction of the samples from the distribution, where the expectation is taken over the probability distribution of the samples.

If $y>1$, then the algorithm chooses to buy at $\sqrt{p_0}$, the adversary can flip the label and cause the $\comp$ (in the worst-case) to become $1 + \frac{1}{\sqrt{p_0}}$ (this happens with probability $p$), and otherwise, the competitive ratio is upper bounded by $(1+\sqrt{p_0})$ (this occurs with probability $\leq 1-p$). 
Hence, in expectation the competitive ratio is therefore $p\left(1+\frac{1}{\sqrt{p_0}}\right) + (1-p)(1+\sqrt{p_0}) < 1 + \sqrt{p_0} + \frac{p}{\sqrt{p_0}}$.

When $y<1$ and $p \leq p_0 \leq \frac{1}{9(e-1)^2}$, we buy at $1$ and our competitive ratio is $2$ with probability $p$ (adversarial) and $1$ with probability $1-p$ (no adversary). Hence, the expected competitive ratio is $1+2p$.
The competitive ratio when the PAC learner is correct is therefore, $\max\{1+2p, 1+\sqrt{p_0} + \frac{p}{\sqrt{p_0}}\} \leq (1+2\sqrt{p_0})$

Now we focus on the points on which the PAC learner makes an error. These comprise $\epsilon$ fraction of the points in the distribution.
When $y$ is predicted to be $\leq 1$ and is actually $>1$, then our competitive ratio is upper bounded by $2$ (since we our always buying before $y$ exceeds 1 and the optimal solution pays $1$). 
When $y$ is predicted to be $>1$ but is actually $y<1$, then the worst case competitive ratio is $1 + \frac{1}{\sqrt{p_0}}$.

We are now ready to calculate the expected competitive ratio as follows:
\begin{align*}
\comp &\leq \left(1+2\sqrt{p_0}\right) \cdot (1 - \epsilon) + \epsilon\cdot\left(1+\frac{1}{\sqrt{p_0}}\right)\\
&\leq 1 + 2(1-\epsilon)\sqrt{p_0}  + \left(\frac{\epsilon}{\sqrt{p_0}}\right)\\
&\leq 1 + 3\sqrt{p_0}.
\end{align*}
% Obviously, if $p_0 > \frac{1}{9(e-1)^2}$, then we get the $\left( 1 + \frac{1}{e-1}\right)$ competitive algorithm by \cite{KMMO94}.
\end{proof}

We also show that the above result is optimal in a rather strong sense: namely,
even with no classification error, the competitive ratio achieved cannot be improved.
\begin{theorem}
\label{thm:pac-noise-lb}
For a given noise rate $p \le \frac{1}{2}$, no (randomized) algorithm can achieve a competitive ratio smaller than $1+\frac{\sqrt{p}}{2}$, even when the algorithm has access to a PAC-learner that has zero classification error.
\end{theorem}
\begin{proof} 
We will show that the adversary can choose a distribution on supplying $y$ that yields a large competitive ratio regardless of the $\theta$ that the algorithm chooses. Let's focus when $y>1$ and the PAC learner correctly predicts this surely.

If there was no adversary, the algorithm should buy at $0$ and $\comp$ is $1$. However the presence of an adversary makes it a bad move, since the adversary can pick $y=\rho$ for an arbitrarily small but positive $\rho$ with a non-zero probability and drive up the competitive ratio arbitrarily.

Here is the exact strategy that the adversary chooses to hurt the algorithm:  the distribution on $y$ is $g(y) = kye^{-y}$. for $y\in[0,\sqrt{p}]$ ($k$ being the normalization constant) This is quite similar to the adversarial distribution chosen in \cite{KMMO94} to enforce an $\frac{e}{e-1}$ ratio.

Now for any value $\theta \in (0,\sqrt{p})$  that the algorithm chooses, the competitive ratio is given by: 
$$\comp(\theta,\mathbb{K}) = p \cdot\left[ \int_{0}^{\theta}g(y)dy + \int_{\theta}^{\sqrt{p}}\frac{(1+\theta)}{y}g(y)dy \right] + (1+\theta)(1-p).$$

Calculating the derivative with respect to $\theta$, we get:
\begin{align*}
\frac{d(\comp(\theta, \mathbb{K}))}{d(\theta)} &= pg(\theta) + p\int_{\theta}^{\sqrt{p}}\frac{g(y)}{y}dy - p\frac{(1+\theta)}{\theta}g(\theta) + (1-p)\\
&= p\frac{g(\theta)}{\theta} + p\int_{\theta}^{\sqrt{p}}ke^{-y}dy + (1-p)\\
&= -pke^{-\theta} + pk(e^{-\theta} - e^{-\sqrt{p}}) + (1-p)\\
&= -pke^{-\sqrt{p}} + (1-p)
\end{align*}

Using the fact that the total probability $\int_{0}^{\sqrt{p}}g(y)dy = 1$ we get that $k = \frac{1}{(1 - (1+\sqrt{p})e^{-\sqrt{p}} )}$. It is easy to see that this value of $k$ gives: $\frac{d(\comp(\theta,\mathbb{K})}{d(\theta)} \leq 0$.

Hence $\comp(\theta,\mathbb{K})$ decreases as $\theta$ goes from 0 to $\sqrt{p}$. Also, the algorithm gains nothing by increasing $\theta$ beyond $\sqrt{p}$. Hence, the best competitive ratio is obtained when algorithm chooses $\theta = \sqrt{p}$. Thus, the algorithm can't hope for a competitive ratio better than \begin{align*}
    \comp(\sqrt{p},\mathbb{K}) &= p\int_{0}^{\sqrt{p}}g(y)dy + (1+\sqrt{p})(1-p)\\
    &=p + (1+\sqrt{p})(1-p)\\
    &=1 + \sqrt{p} - p\sqrt{p} \intertext{For $p<1/2$:}\\
    &\geq 1 + \frac{\sqrt{p}}{2}
\end{align*}
\end{proof}

\section{Robustness Bounds}
In this section, we address the scenario when there is no assumption on the input, i.e., the choice of the input is adversarial. The desirable property in this setting is encapsulated in the following definition of ``robustness'' adapted from the corresponding notion in \cite{purohit2018improving}:
\begin{definition}
A learning-to-rent algorithm $A$ with threshold function $\theta(\cdot)$ is said to be $\gamma$-robust if $g(\theta(x), y) \leq \gamma$ for any feature $x$ and any length of the ski season $y$.
\end{definition}

First, we show an upper bound on the competitive ratio for any algorithm based on the shortest wait time for any input. 
\begin{lemma}\label{lemma:robust}
A learning-to-rent algorithm with threshold function $\theta(\cdot)$ is $\left(1 + \frac{1}{\theta_0}\right)$-robust where:
$$\theta_0 = \min_{x \in \mathbb{R}^{d}}\theta(x).$$
\end{lemma}

\begin{proof}
Note that the function $g(\theta, y)$ achieves its maximum value at $y = \theta + \rho$ where $\rho \rightarrow 0^{+}$. In this case, the algorithm pays $1+\theta$, while the optimal offline cost approaches $\theta$. This gives us that $\max_{y \in \mathbb{R}^{+}}g(\theta,y) = \left(1+\frac{1}{\theta}\right)$. Now, since there is no $x$ such that $\theta(x) < \theta_0$, we get: 
$$\max_{y\in \mathbb{R}^{+}, x \in \mathbb{R}^{d}}g(\theta(x), y) \leq \left(1+\frac{1}{\theta_0}\right)$$. 
\end{proof}

The robustness bounds for our algorithms are straightforward applications of the above lemma.
We derive these bounds below. First, we consider Algorithm~\ref{alg:multi_dim_x} based only on the Lipschitz assumption.

\begin{theorem}
Algorithm~\ref{alg:multi_dim_x} is $\left(1+\frac{1}{\epsilon}\right)$-robust.
\end{theorem}
\begin{proof}
Algorithm~\ref{alg:multi_dim_x} always chooses a threshold in the range $[\epsilon,1/\epsilon]$, i.e., $\theta \ge \epsilon$ for all inputs. The theorem now follows by Lemma~\ref{lemma:robust}.
\end{proof}

Next, we consider the black box algorithm that uses the PAC learning approach, i.e., Algorithm~\ref{algo:naive-pac}.

\eat{
\begin{theorem}
Given access to an $(\eps, \delta)$ PAC-learner that predicts whether the length of the ski season is greater than the buying cost, there exists a learning-to-rent algorithm which is $(1+2\sqrt{\epsilon)}$ competitive with probability $1-\delta$ and has a robustness bound of $\left(1+\frac{1}{\sqrt{\epsilon}}\right)$.
\end{theorem}
\begin{proof}
Given a test input $x$, the algorithm feeds it to the PAC Learner and, sets $\theta(x)$ as:
\begin{equation*}
    \theta(x) = \begin{cases} 1 \, \text{  when PAC learner predicts $y\geq 1$}\\
    \sqrt{\epsilon} \, \text{   otherwise}
    \end{cases}
\end{equation*}
Firstly note that the algorithm has $\theta \ge \sqrt{\epsilon}$ for all inputs, which by Lemma~\ref{lemma:robust} gives us a robustness bound of $1+\frac{1}{\sqrt{\epsilon}}$.
Also, note that when the PAC learner is  correct, we do not incur a competitive ratio of more than $1+\sqrt{\epsilon}$. With probability $1-\delta$. we bound the competitive ratio as:
\begin{align*}
    \comp(\theta) &\leq \left(1 + \frac{1}{\sqrt{\epsilon}}\right)\cdot\epsilon + (1-\epsilon)\cdot(1 + \sqrt{\epsilon})\\
    &\leq 1 + 2\sqrt{\epsilon}.
\end{align*}
\end{proof}
}%end eat

\begin{theorem}
Algorithm~\ref{algo:naive-pac} is $\left(1+\frac{1}{\sqrt{\epsilon}}\right)$-robust.
\end{theorem}
\begin{proof}
Note that Algorithm~\ref{algo:naive-pac} has $\theta \ge \sqrt{\epsilon}$ for all inputs, which by Lemma~\ref{lemma:robust} gives a robustness bound of $1+\frac{1}{\sqrt{\epsilon}}$.
\end{proof}

Next, we show robustness bounds for the margin-based approach, i.e., Algorithm~\ref{algo: margin-based-pac}. 
\eat{
\begin{theorem}
Given a concept class $\mathcal{C}$ with VC-dimension $D(\alpha)$ under margin $\alpha$, there exists a learning-to-rent algorithm that has a competitive ratio of $1 + O(L\alpha)$ for $n$ samples with constant failure probability, where $\alpha$ satisfies:
\begin{equation}
    \label{eq:alpha}
    \sqrt{\frac{D(\alpha)}{n}} = L \alpha.
\end{equation}
Furthermore, this algorithm enjoys a robustness bound of $1+\left(\frac{1}{L\alpha}\right)$.
\end{theorem}
\begin{proof}
The result on the competitive ratio is already shown by Theorem~\ref{thm:margin}.
For the claim on the robustness bound, we note that the minimum $\theta$ set by algorithm~\ref{algo: margin-based-pac} is $\gamma = L\alpha$, which yields a robustness bound of $1+\frac{1}{\gamma} = 1+\left(\frac{1}{L\alpha}\right)$ using Lemma~\ref{lemma:robust}.
\end{proof}
}%end eat
\begin{theorem}
Algorithm~\ref{algo: margin-based-pac} is $\left(1+\frac{1}{L\alpha}\right)$-robust.
\end{theorem}
\begin{proof}
This follows from Lemma~\ref{lemma:robust}, with 
the observation that the shortest wait time 
in Algorithm~\ref{algo: margin-based-pac} is $\gamma = L\alpha$.
\end{proof}

Finally, we consider the noisy classification setting in Algorithm~\ref{alg:noisy}.

\begin{theorem}
Algorithm~\ref{alg:noisy} is 
$\max\left(\frac{e}{e-1}, 1+\frac{1}{\sqrt{\eps}}\right)$-robust.
\end{theorem}
\begin{proof}
In the two cases in Algorithm~\ref{alg:noisy},
either the threshold $\theta$ satisfies 
$\theta\geq \sqrt{p_0}$
or a random threshold is chosen for which the 
expected competitive ratio is $\frac{e}{e-1}$
for any input. In the first, case, we further
note that $p_0 = \max(p, \eps) \geq \eps$,
i.e., $1+\frac{1}{\sqrt{p_0}} \leq 1 + \frac{1}{\sqrt{\eps}}$. 
The theorem now follows by applying 
Lemma~\ref{lemma:robust}.
\end{proof}

%Similar robustness bounds can also be shown for the noise tolerant version of the algorithm (Algorithm~\ref{alg:noisy}) and for General Learning-to-Rent Algorithms by noting their respective minimum wait times, and subsequently applying Lemma~\ref{lemma:robust}.

% \section{Connection to Data Drive Algorithm Design}
% \input data_driven_alg

\section{Numerical Simulations}
\label{sec:experiments}
In this section, we  use numerical simulations to evaluate the algorithms that we designed for the learning-to-rent problem: the black box algorithm (Algorithm~\ref{algo:naive-pac}), the margin-based algorithm (Algorithm~\ref{algo: margin-based-pac}), and the algorithm for a noisy classifier (Algorithm~\ref{alg:noisy}). We compare the first two algorithms and show that as the predicted by the theoretical analysis, the margin-based algorithm substantially outperforms the black box algorithm in high dimensions. For learning-to-rent with a noisy classifier, we show that its competitive ratio follows the $(1+\sqrt{p})$-curve predicted by the theoretical analysis with increasing noise rate $p$.

%In this section, we carry out Experiments to demonstrate two things:
%Firstly, we wish to compare the General PAC Black Box algorithm with the Margin Based approach. The bounds obtained theoretically suggest that the Margin Based method should outperform the general method when the dimension in high and the number of samples is low. We plot the competitive ratio for different sample sizes for the two algorithms.
%We do this for small ($d=2$), large ($d=100$) and very large dimensions ($d=5000$).

\textbf{Experimental Setup.}
We first describe the joint distribution $(x,y)\sim \mathbb{K}$ used in the experiments. 
We choose a random vector $W \in \mathbb{R}^d$ as $W\sim N(0, \mathbf{I}/d)$. We view $W$ as a  hyper-plane passing through the origin $(W^Tx = 0)$. 
The value of $y$, representing the length of the ski season, is calculated as $\frac{2}{(1+e^{-W^{T}x})}$, such that $y\geq 1$ when $W^{T}x\geq 0$ and $y<1$ otherwise. Note that this satisfies the Lipschitz condition given in Definition~\ref{def:lipchitz}, with $L=2$ for $\norm{W} \le 1$. The input $x$ is drawn from a mixture distribution, where with probability $1/2$ we sample $x$ from a Gaussian $x\sim N(0, \mathbf{I}/d)$, and with probability $1/2$, we sample $x$ as $x = \alpha W + \eta$, here $\alpha\sim N(0,1)$ is a coefficient in the direction of $W$ and $\eta\sim N(0, \frac{1}{d}I)$. Choosing $x$ from the Gaussian distribution ensures that the data-set has no margin; however, in high dimensions, $W^Tx$ will concentrate in a small region, which makes all the label $y$ very close to 1. We address this issue by mixing in the second component which ensures that the distribution of $y$ is diverse.

\textbf{Training and Validation.}
For a given training set, we split it in two equal halves, the first half is used to train our PAC learner and the second half is used as a validation set to optimize the design parameters in the algorithms, namely $\tau$ in Algorithm~\ref{algo:naive-pac} and $\gamma$ in Algorithm~\ref{algo: margin-based-pac}. 

%\alert{KEERTI : PARAMETER OPTIMIZATION }

\textbf{Parameter Optimization for Algorithm~\ref{algo:naive-pac} and Algorithm~\ref{algo: margin-based-pac}.}
We perform this optimization on a validation set that is distinct from the training set for these algorithms.

For the black box algorithm (Algorithm~\ref{algo:naive-pac}), we have to choose the value of the parameter $\tau$. Here we set $\tau = c\epsilon$ where $c>0$ is a parameter that we optimize on the validation set. 
In order to do this, we minimize the loss (in this case, the competitive ratio) by running gradient descent from a starting value $c_0$, where $c_0 \in \{1000, 100, 10, 1, 0.1, 0.01\}$. %We pick the value of $c$ that minimizes the competitive ratio.

For the margin based learning-to-rent algorithm (Algorithm~\ref{algo: margin-based-pac}), we optimize the value of $\gamma$ using a similar procedure by running gradient descent from the starting value $\frac{\gamma_0}{N^{1/4}}$, where $\gamma_0 \in \{0.1,0.01,0.001,0.0001,10^{-5}\}$.

 %The distribution $\mathbb{D}$ on the features is chosen as mixture of two distributions. First is a Gaussian with mean $0$ and variance $\frac{1}{d}$ in all orthogonal directions. Second is a distribution along $W$ with some added noise ($x = \alpha \cdot W + \eta$ where $\eta$ is Gaussian noise and $\alpha$ is chosen from uniformly in $\pm 1$). 

\begin{figure*}[!htb]
\begin{multicols}{3}
    \includegraphics[width=\linewidth]{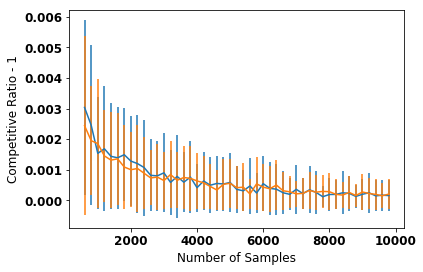}\par 
    \includegraphics[width=\linewidth]{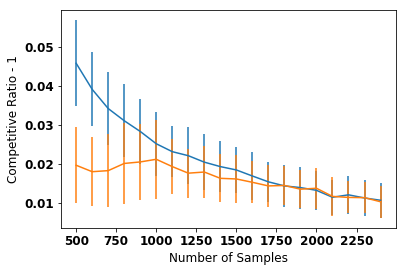}\par 
    \includegraphics[width=\linewidth]{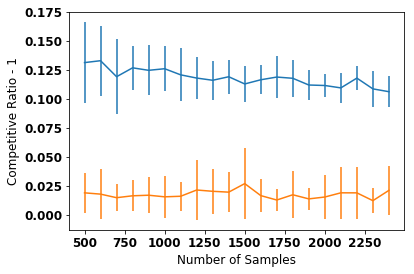}\par
    \end{multicols}
%\begin{multicols}{2}
    % \includegraphics[width=\linewidth]{plots/gmd_5000_err.png}\par
%\end{multicols}
\vspace*{-0.8cm}
\caption{Comparison of Algorithm~\ref{algo:naive-pac} (blue) and Algorithm~\ref{algo: margin-based-pac} (orange). 
From left to right, $d = 2, 100$, and $5000$.}
\label{fig:compare_gen_marg}
\end{figure*}

We test our algorithms for dimensions $d = 2, 100,$ and $5000$. For each $d$, we create a large corpus of samples and select $N$ of them randomly and designate this as the training set; the remaining samples form the test set. 
 
% Following has been sent to appendix 
%  For a given training set, we split it in two equal halves, the first half is used to train our PAC learner and the second half is used as a validation set to optimize the design parameters in the algorithms, namely $\tau$ in Algorithm~\ref{algo:naive-pac} and $\gamma$ in Algorithm~\ref{algo: margin-based-pac}. 
%  (We give details of this parameter optimization procedure in the appendix.) We evaluate the performance of our algorithms as $N$ increases.

\textbf{Comparison between the two algorithms.}
The comparative performance of Algorithm~\ref{algo:naive-pac} and Algorithm~\ref{algo: margin-based-pac} for $d = 2, 100$, and $5000$ is given in Fig.~\ref{fig:compare_gen_marg}.\footnote{In all the figures, the vertical bars represent standard deviation of the output value and the value plotted on the curve is the mean.} 
For small $d$ ($d=2$), we do not see a significant difference in the performance of the two algorithms because the curse of dimensionality suffered by Algorithm~\ref{algo:naive-pac} is not prominent at this stage. In fact, in this case the optimal margin on validation set is very close to 0. % margin based algorithm closely mimicks to the general method (when the margin is optimized to be very thin).
However, as $d$ increases, Algorithm~\ref{algo: margin-based-pac} starts outperforming Algorithm~\ref{algo:naive-pac} as expected from the theoretical analysis. For $d=100$, this difference of performance is prominent at small sample size but disappears for larger samples, because of the trade-off between sample size and number of dimensions in Corollary~\ref{cor:hyperplane} and Theorem~\ref{thm:pac-blackbox}. Eventually, at $d=5000$, Algorithm~\ref{algo: margin-based-pac} is clearly superior.   %a considerable change in the performance of the margin based approach (which allows it to perform better in the presence of fewer samples). 

\eat{
\begin{figure*}[t!]
    \centering
    \begin{subfigure}
        \centering
        \includegraphics[width=0.33\textwidth]{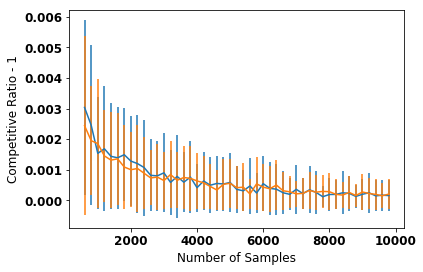}
        \caption{Lorem ipsum}
    \end{subfigure}%
    ~
    \begin{subfigure}
       \centering
        \includegraphics[width=0.33\textwidth]{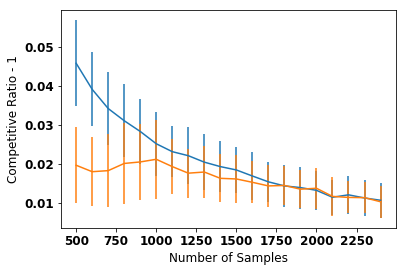}
        \caption{Lorem ipsum, lorem ipsum,Lorem ipsum, lorem ipsum,Lorem ipsum}
    \end{subfigure}%
    ~
    \begin{subfigure}
        \centering
        \includegraphics[width=0.33\textwidth]{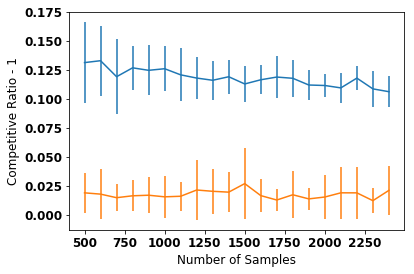}
        \caption{Lorem ipsum, lorem ipsum,Lorem ipsum, lorem ipsum,Lorem ipsum}    
    \end{subfigure}
    \caption{Caption place holder}
\end{figure*}
}

\begin{figure}
    \centering
    \includegraphics[width=0.3\textwidth]{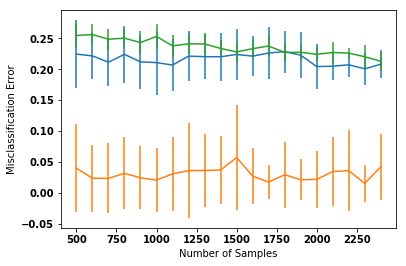}
    \vspace*{-0.4cm}
    \caption{Classification error in Algorithm~\ref{algo:naive-pac} (green) and Algorithm~\ref{algo: margin-based-pac} (blue for all samples, orange for filtered samples).}\label{fig:gmd_5000_err}
    \vspace*{-0.5cm}
\end{figure}

To further understand the difference between the black box approach and the margin-based approach, in Figure~\ref{fig:gmd_5000_err}, we plot the error of the two binary classifiers used in Algorithm~\ref{algo:naive-pac} and Algorithm~\ref{algo: margin-based-pac} when $d = 5000$. Although both classifiers achieve very low accuracy on the entire data-set, the margin-based classifier was able to correctly label the data points that are far from the decision boundary, i.e., the data points where mis-classification would be costly from the optimization perspective. As a result, Algorithm~\ref{algo: margin-based-pac} performs much better overall.

%Note that in figure ~\ref{fig:compare_gen_marg}, we also plot the error of the two binary classification schemes. It is very clear that the margin based PAC learner tries to focus its learning power on the features that are far away from the margin since these samples are more important to make a decision for the algorithm.

\begin{figure}[!htb]
    \centering
    \includegraphics[width=0.3\textwidth]{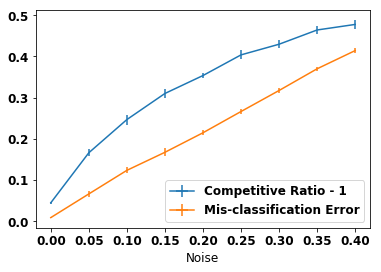}
    \vspace*{-0.5cm}
    \caption{Algorithm~\ref{alg:noisy} with varying noise rate with $d=100$.}
    \label{fig:pac_with_corruption}
\end{figure}

\textbf{Learning with noise.}
We now evaluate the learning-to-rent algorithm with a noisy classifier (Algorithm~\ref{alg:noisy}), We fix the number of dimensions $d=100$, and create a training set of $N=10^5$ samples using the same distribution as earlier. But now, we add noise to the data by declaring each data point as noisy with probability $p$ (we will vary the parameter $p$ over our experiments). There are two types of noisy data points: ones where the classifier predicts $y\ge 1$ and the actual value is $y< 1$, or vice-versa. For data points of the first type, we choose $y$ from the worst case input distribution in the lower bound given by Theorem~\ref{thm:pac-noise-lb}, i.e, 
$\pr[y = z] = \frac{e}{e-1}\cdot z \cdot e^{-z}$ for $z\in [0, 1]$ and point mass of $1/(e-1)$ at some $z > 1$, say at $z=2$. For data points of the second type, the input distribution is not crucial, so we simply choose a uniform random $y$ in $[1,2]$.
The testing is done on a batch of 1000 samples from the same distribution. We use a noise tolerant Perceptron Learner (see, e.g., \cite{bylander1994learning}) to learn the classes ($y\geq 1$ and $y<1$) in the presence of noise. We can see that even for noise rates as high as $40\%$, the competitive ratio of the learning-to-rent algorithm is still better than the $\frac{e}{e-1}$ that is the best achievable in the worst case. (Figure~\ref{fig:pac_with_corruption})

\section{Conclusion and Future Work}
\label{sec:conclusion}
In this paper, we explored the question of customizing 
machine learning algorithms for optimization tasks,
by incorporating optimization objectives in the loss function. We demonstrated, using 
PAC learning, that for the classical rent or buy problem,
the sample complexity of learning can be substantially improved by incorporating the 
insensitivity of the objective to mis-classification near the classification boundary (which is responsible for large sample complexity if accurate classification were the end goal).
In addition, we showed worst-case robustness bounds for our algorithms, i.e., that they exhibit bounded competitive ratios even if the input is adversarial.

This general approach of ``learning for optimization''
opens up a new direction for future research at the boundary of machine learning and 
algorithm design, by providing an alternative ``white box'' approach to the existing
``black box'' approaches for using \ml predictions in {\em beyond worst case} algorithm 
design. While we explored this for an online problem in this paper, the 
principle itself can be applied to any scenario where an algorithm hopes to learn 
patterns in the input that can be exploited to achieve performance gains. 
We posit that this is a rich direction for future research.

%\clearpage

\section*{Acknowledgments}
This work was done under the auspices of the
Indo-US Virtual Networked Joint Center on Algorithms 
under Uncertainty.
K. Anand and D. Panigrahi were supported in part by
NSF grants CCF-1535972, CCF-1955703, 
and an NSF CAREER Award CCF-1750140. 
R. Ge was supported in part by NSF CCF-1704656, 
NSF CCF-1845171 (CAREER), a Sloan Fellowship and a 
Google Faculty Research Award.
\bibliographystyle{unsrt}
\bibliography{references}

% \appendix

% \section{Omitted Details for Section~\ref{sec:lipschitz}: A General Learning-to-Rent Algorithm}
% \label{sec:lipschitz-appendix}
% \input lipschitz-appendix.tex

% \section{Omitted Details for Section~\ref{sec:pac}: A PAC-Learning Approach to the Learning-to-Rent Problem}
% \label{sec:pac-appendix}
% \input pac-appendix.tex

% \input asymmetric-appendix.tex

% \section{Omitted Details for Section~\ref{sec:pac-learning-noise}: Learning-to-Rent with a Noisy Classifier}
% \label{sec:pac-noise-appendix}
% \input pac-noise-appendix.tex

% \section{Notes on Kernels}
% \label{sec:kernel-appendix}
% \input appendix-kernelization.tex

% \section{Details of Experiments}
% \label{sec:experiments-appendix}
% \input experiments-appendix.tex

\end{document}